\pdfoutput=1
\documentclass[10pt,twocolumn,letterpaper]{article}

\usepackage{cvpr}              

\usepackage[utf8]{inputenc} 
\usepackage[T1]{fontenc}    
\usepackage{microtype}      
\usepackage{xcolor,colortbl}         
\usepackage{xspace}
\usepackage{amsmath,amssymb,amsfonts,dsfont,pifont,bm,bbm,mathrsfs,mathtools,nicefrac}
\usepackage{algorithm,algorithmicx,algpseudocode,listings}
\usepackage{wrapfig}
\usepackage{booktabs,multirow,multicol,adjustbox,diagbox,threeparttable}
\usepackage{newfloat,graphicx}

\definecolor{citeblue}{HTML}{0071bc}
\definecolor{textpurple}{RGB}{135,89,201}
\definecolor{Gray}{gray}{0.90}
\newcolumntype{g}{>{\columncolor{Gray}}c}
\definecolor{ffe1da}{RGB}{255,225,218}
\definecolor{F7E0D5}{RGB}{247,224,213}
\definecolor{darkF7E0D5}{RGB}{209,154,128}
\definecolor{Light}{RGB}{255,245,238}

\usepackage[pagebackref=true,breaklinks=true,colorlinks=true,citecolor=citeblue,bookmarks=false]{hyperref}
\usepackage{cleveref}  
\usepackage{bbding}

\newcommand{\method}{MOKD\xspace}
\newcommand{\vm}[1]{\mbox{\boldmath $#1$}}

\newcommand{\supp}{\textit{Supplementary Material}\xspace}

\definecolor{codegreen}{rgb}{0,0.6,0}
\definecolor{codegray}{rgb}{0.5,0.5,0.5}
\definecolor{codepurple}{rgb}{0.58,0,0.82}
\definecolor{backcolour}{rgb}{1.0,1.0,1.0}
\lstdefinestyle{mystyle}{
    backgroundcolor=\color{backcolour},
    commentstyle=\color{codegreen},
    keywordstyle=\color{magenta},
    numberstyle=\tiny\color{codegray},
    stringstyle=\color{codepurple},
    basicstyle=\ttfamily\scriptsize,
    breakatwhitespace=false,
    breaklines=true,
    captionpos=b,
    keepspaces=true,
    showspaces=false,
    showstringspaces=false,
    showtabs=false,
    tabsize=2
}
\crefname{section}{Sec.}{Secs.}
\Crefname{section}{Section}{Sections}
\Crefname{table}{Table}{Tables}
\crefname{table}{Tab.}{Tabs.}
\crefname{figure}{Fig.}{Figs.}
\Crefname{figure}{Figure}{Figures}
\crefname{equation}{Eq.}{Eqs.}

\def\cvprPaperID{1806} 
\def\confName{CVPR}
\def\confYear{2023}

\title{Multi-Mode Online Knowledge Distillation for Self-Supervised Visual Representation Learning}

\author{
    Kaiyou Song\thanks{Corresponding author.} \hspace{0.6mm} \
    Jin Xie \hspace{0.6mm} \
    Shan Zhang \hspace{0.6mm} \
    Zimeng Luo \hspace{0.6mm} \
    \\[3pt]
    Megvii Technology \\[3pt]
    \small{\texttt{\{songkaiyou, xiejin, zhangshan, luozimeng\}@megvii.com}}
}

\begin{document}

\maketitle

\begin{abstract}

Self-supervised learning (SSL) has made remarkable progress in visual representation learning.
Some studies combine SSL with knowledge distillation (SSL-KD) to boost the representation learning performance of small models.
In this study, we propose a Multi-mode Online Knowledge Distillation method (\method) to boost self-supervised visual representation learning.
Different from existing SSL-KD methods that transfer knowledge from a static pre-trained teacher to a student, in \method, two different models learn collaboratively in a self-supervised manner.
Specifically, \method consists of two distillation modes: self-distillation and cross-distillation modes.
Among them, self-distillation performs self-supervised learning for each model independently,
while cross-distillation realizes knowledge interaction between different models.
In cross-distillation, a cross-attention feature search strategy is proposed to enhance the semantic feature alignment between different models.
As a result, the two models can absorb knowledge from each other to boost their representation learning performance.
Extensive experimental results on different backbones and datasets demonstrate that two heterogeneous models can benefit from \method and outperform their independently trained baseline.
In addition, \method also outperforms existing SSL-KD methods for both the student and teacher models.

\end{abstract}
\section{Introduction}
\label{sec:intro}

\begin{figure}[t]
\centering
\includegraphics[width=1.0\columnwidth]{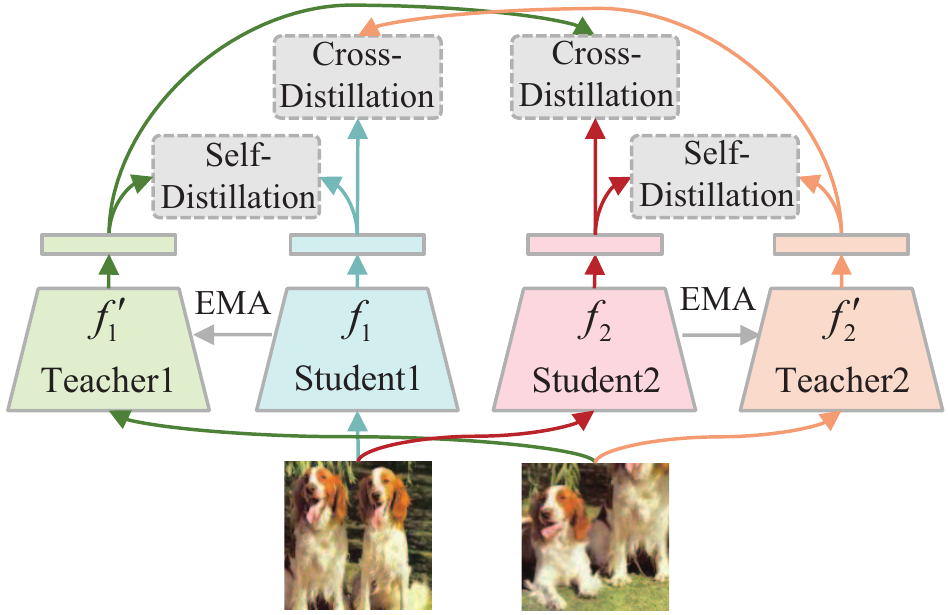}
\caption{Overview of the proposed Multi-mode Online Knowledge Distillation (\method). 
In \method, two different models are trained collaboratively through two types of knowledge distillation modes, i.e., a self-distillation mode and a cross-distillation mode.
EMA denotes exponential-moving-average.}
\label{fig_introduction}
\end{figure}

Due to the promising performance of unsupervised visual representation learning in many computer vision tasks, self-supervised learning (SSL) has attracted widespread attention from the computer vision community.
SSL aims to learn general representations that can be transferred to downstream tasks by utilizing massive unlabeled data.

Among various SSL methods, contrastive learning \cite{moco_2020,simclr_2020} has shown significant progress in closing the performance gap with supervised methods in recent years. 
It aims at maximizing the similarity between views from the same instance (positive pairs) while minimizing the similarity among views from different instances (negative pairs).
MoCo \cite{moco_2020,mocov2_2020} and SimCLR \cite{simclr_2020,simclrv2_2020} use both positive and negative pairs for contrast.
They significantly improve the performance compared to previous methods \cite{cpc_2018,insdis_2018}.
After that, many methods are proposed to solve the limitations in contrastive learning, such as the false negative problem \cite{msf_2021,isd_2021,nnclr_2021,univip_2022,hcsc_2022}, the limitation of large batch size \cite{momentum2tea_2021, dcl_2021}, and the problem of hard augmented samples \cite{clsa_2021,adco_2021}.
At the same time, other studies \cite{byol_2020,simsiam_2021,swav_2020,barlowtwins_2021,vicreg_2021,obow_2021,dino_2021} abandon the negative samples during contrastive learning.
With relatively large models, such as ResNet50 \cite{resnet_2016} or larger, these methods achieve comparable performance on different tasks than their supervised counterparts.
However, as revealed in previous studies \cite{seed_2021, disco_2022}, they do not perform well on small models \cite{mobilenet_v3_2019,efficientnet_2019} and have a large gap from their supervised counterparts.

To address this challenge in contrastive learning, some studies \cite{crd_2020, simclrv2_2020, compress_2020, simreg_2021, seed_2021, rekd_2022, disco_2022} propose to combine knowledge distillation \cite{kd_2015} with contrastive learning (SSL-KD) to improve the performance of small models.
These methods first train a larger model in a self-supervised manner and then distill the knowledge of the trained teacher model to a smaller student model.
There is a limitation in these SSL-KD methods, i.e., knowledge is distilled to the student model from the static teacher model in a unidirectional way.
The teacher model cannot absorb knowledge from the student model to boost its performance.

In this study, we propose a Multi-mode Online Knowledge Distillation method (\method), as illustrated in \cref{fig_introduction}, to boost the representation learning performance of two models simultaneously.
Different from existing SSL-KD methods that transfer knowledge from a static pre-trained teacher to a student, in \method, two different models learn collaboratively in a self-supervised manner.
Specifically, \method consists of a self-distillation mode and a cross-distillation mode.
Among them, self-distillation performs self-supervised learning for each model independently,
while cross-distillation realizes knowledge interaction between different models.
In addition, a cross-attention feature search strategy is proposed in cross-distillation to enhance the semantic feature alignment between different models.
Extensive experimental results on different backbones and datasets demonstrate that model pairs can both benefit from \method and outperform their independently trained baseline.
For example, when trained with ResNet \cite{resnet_2016} and ViT \cite{vit_2021}, two models can absorb knowledge from each other, and representations of the two models show the characteristics of each other.
In addition, \method also outperforms existing SSL-KD methods for both the student and teacher models.

The contributions of this study are threefold:
\begin{itemize}
\item We propose a novel self-supervised online knowledge distillation method, i.e., \method.
\item \method can boost the performance of two models simultaneously, achieving state-of-the-art contrastive learning performance on different models.
\item \method achieves state-of-the-art SSL-KD performance.
\end{itemize}
\section{Related Works}
\label{sec:relatedwork}

\subsection{Knowledge Distillation}
\label{subsec:rw_kd}

Knowledge distillation \cite{kd_2015} aims to distill knowledge from a larger teacher model to a smaller student model to improve the performance of the student model.
Many studies have been proposed in recent years, which can be divided into three groups, i.e., logits-based, feature-based, and relation-based, according to the knowledge types.

Logits-based \cite{kd_2015, improved_kd_2020} knowledge distillation utilizes the logits of the teacher model as the knowledge.
In the vanilla knowledge distillation \cite{kd_2015}, the student model mimics the logits of the teacher model by minimizing the KL-divergence of the class distribution.
Feature-based methods \cite{fitnets_2014,pkt_2018,SemCKD_2021} utilize the output of intermediate layers, i.e., feature maps, as the knowledge to supervise the training of the student model.
Relation-based knowledge distillation \cite{rkd_2019,crcd_2021} distills the relation between samples rather than a single instance.

These methods mentioned above perform offline distillation.
Some studies \cite{dml_2018, onlinekd_zhu_2018, onlinekd_guo_2020, onlinekd_chen_2020, onlinekd_zhang_2021} are developed to perform online distillation, i.e., the teacher and the student model are trained simultaneously.
Deep mutual learning \cite{dml_2018} is first proposed to train multiple models collaboratively.
After that, studies are proposed to improve deep mutual learning regarding generalization ability \cite{onlinekd_guo_2020, onlinekd_chen_2020} and computation efficiency \cite{onlinekd_zhu_2018}.
All these methods are trained in a supervised manner.

\subsection{Self-Supervised Knowledge Distillation}
\label{subsec:rw_sslkd}

Due to significant improvement for small models, knowledge distillation is introduced to self-supervised learning to improve the performance of small models.
CRD \cite{crd_2020} combines a contrastive loss with knowledge distillation to transfer the structural knowledge of the teacher model.
SimCLR-v2 \cite{simclrv2_2020} proposes to train a larger model via self-supervised learning first and uses the supervised finetuned large model to distill a smaller model via self-supervised learning.
SSKD \cite{sskd_2020} combines self-supervised learning with supervised learning to transfer richer knowledge.
Compress \cite{compress_2020} and SEED \cite{seed_2021} transfer the knowledge of probability distribution in a self-supervised manner by utilizing the memory bank in MoCo \cite{moco_2020}.
SimReg \cite{simreg_2021} directly conducts feature distillation by minimizing the squared Euclidean distance between the features of the teacher and student.
While ReKD \cite{rekd_2022} transfers the relation knowledge to the student.
DisCo \cite{disco_2022} proposes to transfer the final embeddings of a self-supervised pre-trained teacher.
There is a limitation in these SSL-KD methods, i.e., knowledge is distilled to a student model from a static teacher model in a unidirectional way.
The teacher model cannot absorb knowledge from the student model.
Recently, DoGo \cite{dogo_2021} and MCL \cite{mcl_2022} combined MoCo \cite{moco_2020} with mutual learning \cite{dml_2018} for online SSL-KD.
However, they either lack a direct comparison with SSL-KD methods on mainstream backbones and tasks or can't guarantee the performance of larger models.
\section{Methods}
\label{sec:methods}

In this section, we first introduce the overall architecture of \method in \cref{subsec_method_overall}.
Then, the two distillation modes of \method, i.e., self-distillation and cross-distillation, are introduced in \cref{subsec_method_selfdis} and \cref{subsec_method_crossdis}, respectively.
Finally, the training procedure and implementation details are introduced in \cref{subsec_method_implementation}.

\begin{figure}[t]
\centering
\includegraphics[width=1.0\columnwidth]{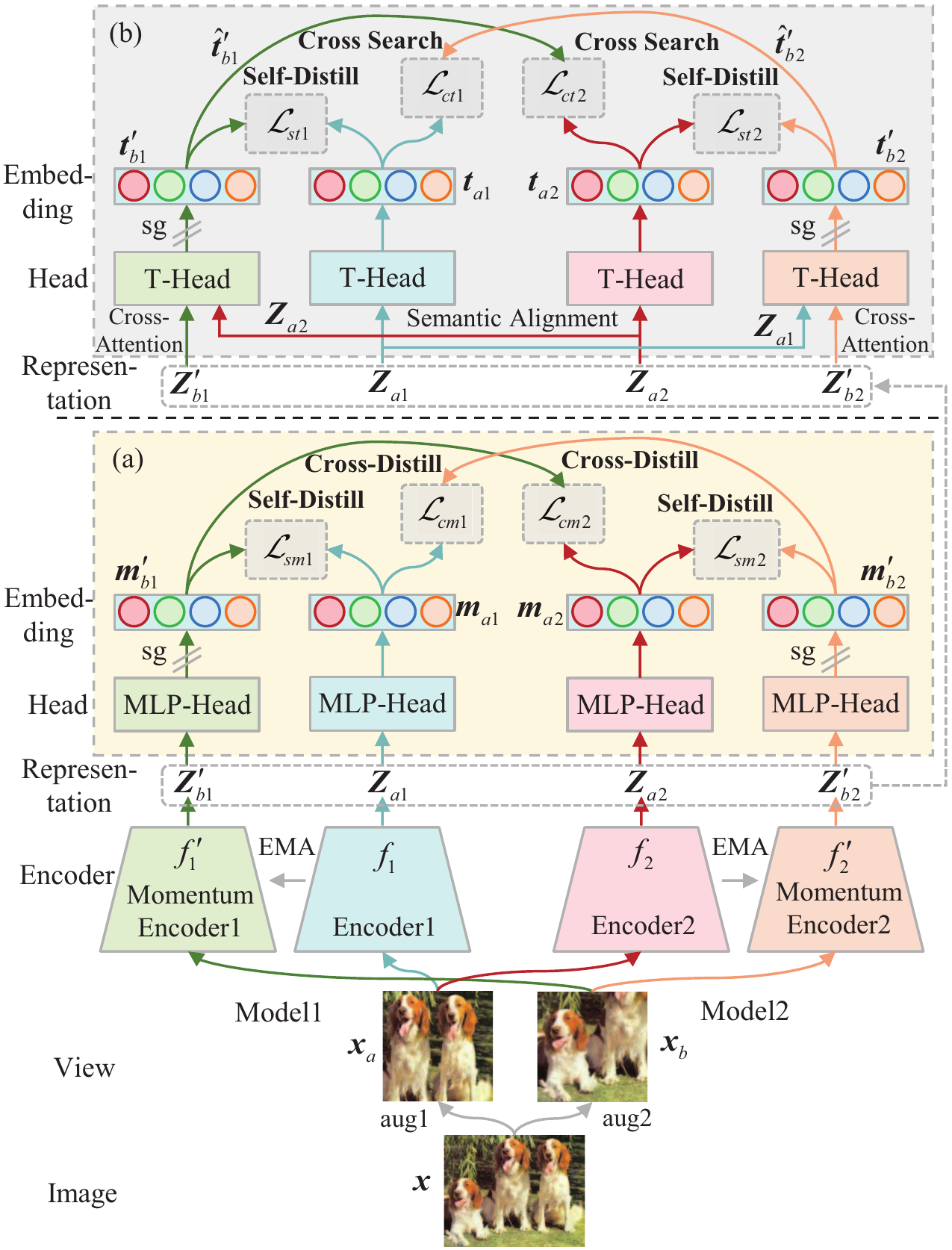}
\caption{
The overall architecture of \method.
In \method, two different models (model1 and model2) are trained collaboratively in a self-supervised manner.
There are two types of knowledge distillation modes: self-distillation and cross-distillation modes.
And the distillation procedure is performed in two feature spaces projected by two types of projection heads, i.e., (a) MLP-Head and (b) T-Head.
``sg" denotes the stop-gradient operation.
}
\label{fig_method_MOKD}
\end{figure}

\subsection{Overall Architecture}
\label{subsec_method_overall}

The overall architecture of \method is shown in \cref{fig_method_MOKD}.
In \method, two different models $\vm{f}_i$ ($i=1,2$) are trained collaboratively in a self-supervised manner.
There are two knowledge distillation modes: self-distillation and cross-distillation modes.
In each model, a multi-layer-perceptron head (MLP-Head) (\cref{fig_method_MOKD}(a)) and a Transformer head (T-Head) (\cref{fig_method_MOKD}(b)) are employed to project the feature representations $\vm{Z}$ produced by the encoders to the output embeddings $\vm{m}$ and $\vm{t}$ for self-distillation and cross-distillation.
Here, the T-Head, which consists of several Transformer blocks, is designed to enhance the semantic alignment between the two models.
Self-distillation, which is conducted between each model $\vm{f}_i$ (as a student) and its EMA version model $\vm{f}'_i$ (as a teacher), performs self-supervised learning for each model independently.
The self-distillation losses are ${\mathcal{L}_{smi}}$ and ${\mathcal{L}_{sti}}$ for the MLP-Head and T-Head, respectively, which will be introduced in \cref{subsec_method_selfdis}.
While cross-distillation, which is conducted between the two models, is employed for knowledge interaction between the two models.
In cross-distillation, by utilizing the self-attention mechanism of the T-Head, we design a cross-attention feature search strategy to enhance semantic alignment between different models.
The cross-distillation losses are ${\mathcal{L}_{cmi}}$ and ${\mathcal{L}_{cti}}$ for the MLP-Head and T-Head, respectively, which will be introduced in \cref{subsec_method_crossdis}.
Here, the subscript $s$ and $c$ stand for self-distillation and cross-distillation, respectively.
And the subscript $m$ and $t$ stand for MLP-Head and T-Head, respectively.

\subsection{Self-Distillation}
\label{subsec_method_selfdis}

Self-distillation performs the contrastive learning task for each model independently.
In this study, we design self-distillation based on the contrastive learning method DINO \cite{dino_2021}.
Specifically, take the model1 as an example.
Given two augmentations ($\vm{x}_a$ and $\vm{x}_b$) of an input image $\vm{x}$, the backbone encoder $f_1$ and its EMA version (the momentum encoder $f'_1$) encode them into the representations: $\vm{Z}_{a1}\!=\!f_1(\vm{x}_a)$, $\vm{Z}'_{b1}\!=\!f'_1(\vm{x}_b)$.
The representations are the feature maps (for convolution neural networks (CNN) \cite{resnet_2016}) or tokens (for vision transformers \cite{vit_2021}) before global average pooling.
Then, the representations are globally-average-pooled and fed into the corresponding MLP-Head to obtain the final embeddings $\vm{m}_{a1}$ and $\vm{m}'_{b1}$.
$\vm{m}_{a1}$, $\vm{m}'_{b1} \in {\mathbb{R}^K}$, $K$ is the output dimension.
The embeddings are normalized with a softmax function:
\begin{equation}
\vm{m}_{a1}^i = \frac{{\exp \left( {{{\vm{m}_{a1}^i} \mathord{\left/
 {\vphantom {{\vm{m}_{a1}^i} \tau }} \right.
 \kern-\nulldelimiterspace} \tau }} \right)}}{{\sum\nolimits_{k = 1}^K {\exp \left( {{{\vm{m}_{a1}^k} \mathord{\left/
 {\vphantom {{\vm{m}_{a1}^k} \tau }} \right.
 \kern-\nulldelimiterspace} \tau }} \right)} }}
\label{equ:l_softmax}
\end{equation}
where $\tau>0$ is a temperature parameter that controls the sharpness of the output distribution.
Note that $\vm{m}'_{b1}$ is also normalized with a similar softmax function with temperature $\tau'$.
$\vm{x}_a$ and $\vm{x}_b$ are fed to the momentum encoder and encoder symmetrically and $\vm{m}'_{a1}$ and $\vm{m}_{b1}$ are obtained respectively.
Following DINO \cite{dino_2021}, the cross-entropy loss is employed as the contrastive loss.
This task is a dynamic self-distillation procedure where student (encoder) and teacher (momentum encoder) have the same architecture.
Similar self-distillation loss can be calculated for the model2, as follows:
\begin{equation}
\left\{ \begin{array}{l}
{\mathcal{L}_{sm1}} =  -\frac{1}{2}\left( {\vm{m}'_{b1}}\log \left( {\vm{m}_{a1}} \right) + {\vm{m}'_{a1}}\log \left( {\vm{m}_{b1}} \right) \right)\\
{\mathcal{L}_{sm2}} =  - \frac{1}{2}\left( {\vm{m}'_{b2}}\log \left( {\vm{m}_{a2}} \right) + {\vm{m}'_{a2}}\log \left( {\vm{m}_{b2}} \right) \right)
\end{array} \right.
\label{equ:l_sm}
\end{equation}
Following DINO \cite{dino_2021}, we also employ the same whitening strategy to avoid model collapse and multi-crop \cite{swav_2020} to enrich augmentations.

As shown in \cref{fig_method_MOKD}(b), self-distillation is also conducted on the output embeddings $\vm{t}$ of T-Head to stabilize the training of T-Head.
A detailed explanation will be introduced in \cref{subsec_method_crossdis}.
Specifically, the representations $\vm{Z}$ are fed into the corresponding T-Head to obtain the final embeddings $\vm{t}$.
$\vm{t} \in {\mathbb{R}^K}$, $K$ is the output dimension.
After the same softmax operation in \cref{equ:l_softmax}, the self-distillation loss of T-Head is calculated:
\begin{equation}
\left\{ \begin{array}{l}
{\mathcal{L}_{st1}} =  - \frac{1}{2}\left( {{{\vm{t}'}_{b1}}\log \left( {{\vm{t}_{a1}}} \right) + {{\vm{t}'}_{a1}}\log \left( {{\vm{t}_{b1}}} \right)} \right)\\
{\mathcal{L}_{st2}} =  - \frac{1}{2}\left( {{{\vm{t}'}_{b2}}\log \left( {{\vm{t}_{a2}}} \right) + {{\vm{t}'}_{a2}}\log \left( {{\vm{t}_{b2}}} \right)} \right)
\end{array} \right.
\label{equ:l_st}
\end{equation}

The self-distillation loss for each model is the sum of the self-distillation losses of MLP-Head and T-Head:
\begin{equation}
\left\{ \begin{array}{l}
{\mathcal{L}_{self1}} = {\mathcal{L}_{sm1}} + {\mathcal{L}_{st1}}\\
{\mathcal{L}_{self2}} = {\mathcal{L}_{sm2}} + {\mathcal{L}_{st2}}
\end{array} \right.
\label{equ:l_self}
\end{equation}

\subsection{Cross-Distillation}
\label{subsec_method_crossdis}

Cross-distillation realizes the interactive learning between two models.
We design two interactive learning objectives, i.e., cross-distillation using MLP-Head embedding and cross-distillation using T-Head embedding, to realize the knowledge transfer between two models.

For MLP-Head embedding, it contains rich knowledge of each model.
Thus, cross-distillation is conducted between two models to interact knowledge.
Specifically, model1 learns knowledge from the momentum version of model2 and vice versa.
The cross-distillation can be calculated as follows:
\begin{equation}
\left\{ \begin{array}{l}
{\mathcal{L}_{cm1}}\! =\!  - \frac{1}{2}\left( {{{\vm{m}'}_{b2}}\log \left( {{\vm{m}_{a1}}} \right)\! +\! {{\vm{m}'}_{a2}}\log \left( {{\vm{m}_{b1}}} \right)} \right)\\
{\mathcal{L}_{cm2}}\! =\!  - \frac{1}{2}\left( {{{\vm{m}'}_{b1}}\log \left( {{\vm{m}_{a2}}} \right)\! +\! {{\vm{m}'}_{a1}}\log \left( {{\vm{m}_{b2}}} \right)} \right)
\end{array} \right.
\label{equ:l_cm}
\end{equation}
Cross-distillation is conducted between different views and different models (online and another momentum model), which has two advantages.
First, cross-distillation between different views can relax the constraint for the same view and is helpful for avoiding the homogenization of two models.
Second, cross-distillation between an online model and another momentum model, rather than two online models, provides more stable training since the momentum model is more stable.

\begin{figure}[t]
\centering
\includegraphics[width=0.8\columnwidth]{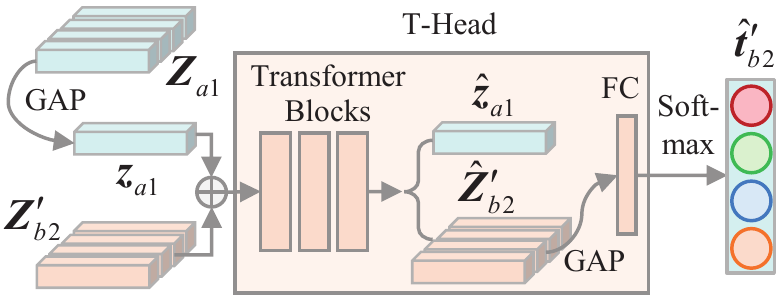}
\caption{Cross-attention feature search in T-Head. $\oplus$ denotes the concat operation.}
\label{fig_method_ca}
\end{figure}

During cross-distillation, the knowledge transfer of semantics-relevant features from different views should be enhanced while irrelevant features should be suppressed.
To this end, the cross-attention feature search is proposed to search semantics-relevant features from each other for knowledge transfer adaptively.
As shown in \cref{fig_method_ca}, the T-Head is designed to apply the self-attention mechanism in Transformer \cite{transformer_2017} to realize the feature search.

Take the cross-attention feature search between model1 and momentum model2 as an example.
We aim to search semantics-relevant features between the feature $\vm{Z}_{a1}$ (${\vm{Z}_{a1}}\!\in\! {\mathbb{R}^{{N_1} \times {C_1}}}$, $N_1$ denotes the number of local features, which is the product of width and height of feature map for CNN and token number for vision transformer, and $C_1$ denotes the dimension of local features) of encoder1 and the feature $\vm{Z}'_{b2}$ (${\vm{Z}'_{b2}}\!\in\! {\mathbb{R}^{{N_2} \times {C_2}}}$, $N_2$ and $C_2$ denote similar information for encoder2) of momentum encoder2.
A global average pooling and a $1\!\times\!1$ convolution operation are conducted on $\vm{Z}_{a1}$ to obtain its global feature and unify its dimension with $\vm{Z}'_{b2}$.
Then, the obtained ${\vm{z}_{a1}}$ (${z_{a1}}\!\in\!{\mathbb{R}^{{C_2}}}$) is concatenated with $\vm{Z}'_{b2}$ and fed to T-Head:
\begin{equation}
\left[ {{{\hat {\vm{z}}}_{a1}},{{\hat {\vm{Z}}'}_{b2}}} \right] = {{f'}_{t2}}\left( {\left[ {{\vm{z}_{a1}},{{\vm{Z}'}_{b2}}} \right]} \right)
\label{equ:l_thead}
\end{equation}
where ${{f'}_{t2}}\left(  \cdot  \right)$ denotes the function of the transformer blocks in T-Head of momentum model2, and $[ \cdot , \cdot ]$ refers to the concatenation operation.
Through the self-attention mechanism in T-Head, the obtained feature ${\hat {\vm{Z}}'}_{b2}$ enhances the semantics consistent component while suppressing the irrelevant component with ${\vm{z}_{a1}}$.
After a global average pooling, an FC layer, and softmax in T-Head, the output embedding ${{{\hat {\vm{Z}}'}_{b2}}}$ is used for contrast learning with the embedding ${\vm{t}_{a1}}$ of ${\vm{Z}_{a1}}$. 
A similar cross-attention feature search procedure is conducted between model2 and momentum model1.
The loss for cross-attention feature search can be calculated:
\begin{equation}
\left\{ \begin{array}{l}
{\mathcal{L}_{ct1}} =  - \frac{1}{2}( {{{\hat {\vm{t}}'}_{b2}}\log \left( {{\vm{t}_{a1}}} \right) + {{\hat {\vm{t}}'}_{a2}}\log \left( {{\vm{t}_{b1}}} \right)})\\
{\mathcal{L}_{ct2}} =  - \frac{1}{2}( {{{\hat {\vm{t}}'}_{b1}}\log \left( {{\vm{t}_{a2}}} \right) + {{\hat {\vm{t}}'}_{a1}}\log \left( {{\vm{t}_{b2}}} \right)} )
\end{array} \right.
\label{equ:loss_ct}
\end{equation}

The T-Head of the momentum model cannot be updated if there is only the feature search loss for T-Head.
Therefore, self-distillation is also conducted between T-Head embeddings (\cref{equ:l_st}) to enable the updating of T-Head and provide more stable training.
The cross-distillation loss for each model is the sum of the contrastive losses of the MLP-Head and T-Head:
\begin{equation}
\left\{ \begin{array}{l}
{\mathcal{L}_{cross1}} = {\mathcal{L}_{cm1}} + {\mathcal{L}_{ct1}}\\
{\mathcal{L}_{cross2}} = {\mathcal{L}_{cm2}} + {\mathcal{L}_{ct2}}
\end{array} \right.
\label{equ:l_cross}
\end{equation}

The overall loss for each model is the weighted sum of self-distillation loss and cross-distillation loss:
\begin{equation}
\left\{ \begin{array}{l}
{\mathcal{L}_1} = {\mathcal{L}_{self1}} + {\lambda _1}{\mathcal{L}_{cross1}}\\
{\mathcal{L}_2} = {\mathcal{L}_{self2}} + {\lambda _2}{\mathcal{L}_{cross2}}
\end{array} \right.
\label{equ:l_all}
\end{equation}
where $0\! \le\! {\lambda _1}\! \le\! 1$ and $0\! \le\! {\lambda _2}\! \le\! 1$ are hyper-parameters and denote the weight of the cross-distillation loss of model1 and model2, respectively.

\subsection{Implementation Details}
\label{subsec_method_implementation}
\noindent
\textbf{Training Procedure.}
In \method, two models are trained cooperatively.
Algorithm \ref{algo:MOKD} summarizes the training procedure of \method.
The SGD and AdamW \cite{adamw_2018} optimizers are used for CNN and ViT, respectively.

\noindent
\textbf{Projection Head.}
MLP-Head consists of a four-layer MLP with the same architecture as DINO~\cite{dino_2021}.
T-Head consists of 3 transformer blocks with the same architecture as ViT-Small \cite{vit_2021} and an FC layer for projection.
The output dimension of the two heads is $K\!=\!65536$.

\begin{algorithm}[t]
\caption{PyTorch-Style Pseudocode of \method.}
\label{algo:MOKD}
\begin{lstlisting}[language=python]
# n1, n1_, n2, n2_: nets and momentum nets
# t1_, t2_: T-Heads of momentum nets
# l: network momentum rates
n1_.params, n2_.params = n1.params, n2.params
for x in loader: # load a minibatch x with n samples
    xa, xb = augment(x), augment(x) # random views
    # net1 and momentum net1 output
    [ma1,ta1,Za1], [mb1,tb1,Zb1] = n1(xa), n1(xb)
    [ma1_,ta1_,Za1_], [mb1_,tb1_,Zb1_] = n1_(xa),n1_(xb)
    
    # net2 and momentum net2 output
    [ma2,ta2,Za2], [mb2,tb2,Zb2] = n2(xa), n2(xb)
    [ma2_,ta2_,Za2_], [mb2_,tb2_,Zb2_] = n2_(xa),n2_(xb)
    
    # cross-attention feature search
    tb1_s, ta1_s = t1_(Za2, Zb1_), t1_(Zb2, Za1_)
    tb2_s, ta2_s = t2_(Za1, Zb2_), t2_(Zb1, Za2_)
    
    # self-distillation loss
    loss_sm1 = H(mb1_,ma1)/2 + H(ma1_,mb1)/2
    loss_sm2 = H(mb2_,ma2)/2 + H(ma2_,mb2)/2
    loss_st1 = H(tb1_,ta1)/2 + H(ta1_,tb1)/2
    loss_st2 = H(tb2_,ta2)/2 + H(ta2_,tb2)/2
    
    # cross-distillation loss
    loss_cm1 = H(mb2_,ma1)/2 + H(ma2_,mb1)/2
    loss_cm2 = H(mb1_,ma2)/2 + H(ma1_,mb2)/2
    loss_it1 = H(tb2_s,ta1)/2 + H(ta2_s,tb1)/2
    loss_it2 = H(tb1_s,ta2)/2 + H(ta1_s,tb2)/2
    
    # total loss
    loss1 = (loss_sm1 + loss_st1) + lamda1 * (loss_cm1 + loss_ct1)
    loss2 = (loss_sm2 + loss_st2) + lamda2 * (loss_cm2 + loss_ct2)
    
    # back-propagate
    loss1.backward(), loss2.backward()
    
    # nets update
    update(n1), update(n2)
    n1_.params = l*n1.params + (1-l)*n1.params
    n2_.params = l*n2.params + (1-l)*n2.params
    
def H(t, s):
    return - (t * log(s)).sum(dim=1).mean()
\end{lstlisting}
\end{algorithm}
\section{Experiments}
\label{sec:experiments}
 
In this section, we conduct comprehensive experiments to evaluate the effectiveness of \method.
Different sizes of CNNs and vision transformers are used as encoders.
Heterogeneous and homogeneous models are evaluated.
For heterogeneous \method, a ResNet \cite{resnet_2016} and a ViT \cite{vit_2021} are used.
Specifically, ResNet101 (R101)/ResNet50 (R50) is used for CNN, and ViT-Base (ViT-B)/ViT-Small (ViT-S) is used for vision transformer.
For homogeneous \method, two CNNs or two ViTs are used.
For two CNNs, R101/R50 is used for the larger model, ResNet34 (R34)/ResNet18 (R18) is used for the smaller model.
For two ViTs, ViT-S and ViT-Tiny (ViT-T) are used.
In addition, we also conduct experiments for two models have the same architecture, including R50, R18, and ViT-S.
All models are trained on the ImageNet \cite{imagenet_2015} training set.
Without a specific statement, the default setting is 256 batch size and 100 epochs.
We follow most hyper-parameters settings of DINO \cite{dino_2021}.
More details for experiments can be found in \supp.

\subsection{Experiments on ImageNet}
\label{subsec_exp_imagenets}
After pre-training, the $k$-NN and linear probing are employed to evaluate the representation performance.
For linear probing, a linear classifier added to the frozen backbone is trained for 100 epochs \cite{moco_2020}.
The top-1 accuracy on the validation set is adopted as the evaluation metric.

\noindent
\textbf{$k$-NN and Linear Probing Accuracy.}
\method is compared with the baseline which two models are trained independently using DINO \cite{dino_2021}.
The results are shown in \cref{tab_knnprob}.
For heterogeneous models (ResNet-ViT), \method significantly improves the performance of two models compared with models that are trained independently.
For example, with R50-ViT-B, the linear probing accuracy of the two models improves by 3.5\% (from 72.1\% to 75.6\%) and 1.0\% (from 77.0\% to 78.0\%), respectively.
With homogeneous models (two ResNets and two ViTs), \method improves the performance of the smaller model with a large margin.
The experimental results demonstrate that \method can effectively transfer knowledge between different models to boost the representation performance.

\begin{table}[]
\centering
\setlength{\abovecaptionskip}{0.2cm}
\setlength{\tabcolsep}{0.3mm}{
\begin{tabular}{ll|cc|>{\columncolor{Light}}c>{\columncolor{Light}}c|cc|>{\columncolor{Light}}c>{\columncolor{Light}}c}
\toprule
\multicolumn{2}{c}{}\vline & \multicolumn{4}{c}{$k$-NN}\vline & \multicolumn{4}{c}{Linear Probing} \\ \hline
\multicolumn{2}{c}{Backbone} \vline & \multicolumn{2}{c}{Independent} \vline & \multicolumn{2}{c}{\method}\vline & \multicolumn{2}{c}{Independent}\vline & \multicolumn{2}{c}{\method} \\ \hline
Net1 & Net2 & Net1 & Net2 & Net1 & Net2 & Net1 & Net2 & Net1 & Net2 \\ \hline
R50 & ViT-S & 62.8 & 69.9 & \textbf{67.1} & \textbf{70.8} & 72.1 & 73.8 & \textbf{74.1} & \textbf{74.4} \\
R50 & ViT-B & 62.8 & 73.6 & \textbf{70.6} & \textbf{75.2} & 72.1 & 77.0 & \textbf{75.6} & \textbf{78.0} \\
R101 & ViT-S & 66.9 & 69.9 & \textbf{68.3} & \textbf{70.7} & 74.6 & 73.8 & \textbf{75.0} & \textbf{74.7} \\
R101 & ViT-B & 66.9 & 73.6 & \textbf{68.5} & \textbf{74.7} & 74.6 & 77.0 & \textbf{75.2} & \textbf{77.5} \\ \hline
R50 & R34 & 62.8 & 60.6 & \textbf{62.9} & \textbf{61.2} & 72.1 & 66.5 & \textbf{72.3} & \textbf{67.0} \\
R50 & R18 & 62.8 & 53.2 & 62.7 & \textbf{57.1} & 72.1 & 61.2 & 72.0 & \textbf{63.4} \\
R101 & R34 & 66.9 & 60.6 & 66.9 & \textbf{61.7} & 74.6 & 66.5 & \textbf{74.7} & \textbf{67.6} \\
R101 & R18 & 66.9 & 53.2 & 66.7 & \textbf{57.4} & 74.6 & 61.2 & 74.5 & \textbf{63.6} \\
R50 & R50 & 62.8 & 62.8 & \textbf{63.0} & \textbf{63.1} & 72.1 & 72.1 & \textbf{72.4} & \textbf{72.4} \\
R18 & R18 & 53.2 & 53.2 & \textbf{53.4} & \textbf{53.4} & 61.2 & 61.2 & \textbf{61.3} & \textbf{61.4} \\ \hline
ViT-S & ViT-T & 69.9 & 59.9 & \textbf{70.6} & \textbf{62.1} & 73.8 & 63.8 & \textbf{74.2} & \textbf{64.3} \\
ViT-S & ViT-S & 69.9 & 69.9 & \textbf{70.5} & \textbf{70.4} & 73.8 & 73.8 & \textbf{74.3} & \textbf{74.2} \\
\bottomrule
\end{tabular}}
\caption{$k$-NN and linear probing accuracy (\%) on ImageNet.}
\label{tab_knnprob}
\end{table}

\noindent
\textbf{Compared with SSL Methods.}
The performance of \method for different models is compared with other outstanding SSL methods.
Most contrastive learning methods conduct experiments using R50, and several methods use ViT-S and ViT-B.
Thus, we compare the performance of the three models.
As shown in \cref{tab_compared_ssl}, \method achieves the best performance for R50, ViT-S, and ViT-B.

\begin{table}[t]
\centering
\setlength{\abovecaptionskip}{0.2cm}
\setlength{\tabcolsep}{0.3mm}{
\begin{tabular}{l|lllcl}
\toprule
Method        & Backbone & BS & Epoch & $k$-NN  & LP   \\ \hline
SimCLR~\cite{simclr_2020}        & R50      & 4096       & 1000  & -    & 69.3          \\
BYOL~\cite{byol_2020}          & R50      & 4096       & 1000  & 66.9 & 74.3          \\
SwAV~\cite{swav_2020}          & R50      & 4096       & 800   & -    & 75.3          \\
MoCo-v2~\cite{mocov2_2020}       & R50      & 256        & 200   & 55.6 & 67.5          \\
SimSiam~\cite{simsiam_2021}       & R50      & 256        & 200   & -    & 70.0          \\
MSF~\cite{msf_2021}           & R50      & 256        & 200   & 64.9 & 72.4          \\
NNCLR~\cite{nnclr_2021}         & R50      & 4096       & 200   & -    & 70.7          \\
Triplet~\cite{triplet_2021}       & R50      & 832        & 200   &  -    & 74.1          \\
Barlow Twins~\cite{barlowtwins_2021}  & R50      & 2048       & 1000  &   -   & 73.2  \\
OBoW~\cite{obow_2021}          & R50      & 256        & 200   & -    & 73.8          \\
AdCo~\cite{adco_2021}          & R50      & 256        & 200   & -    & 73.2          \\
MoCo-v3~\cite{mocov3_2021}       & R50      & 4096       & 300   & -    & 72.8          \\
UniVIP~\cite{univip_2022}       & R50      & 4096       & 200   & -    & 73.1          \\
HCSC~\cite{hcsc_2022}       & R50      & 256       & 200   & -    & 73.3          \\
DINO~\cite{dino_2021}          & R50      & 4080       & 800   & 67.5 & 75.3          \\
\rowcolor{Light} \textbf{\method} (R50-ViT-B)  & R50      & 256        & 100   & \textbf{70.6} & \textbf{75.6} \\ \hline
SimCLR~\cite{simclr_2020}        & ViT-S    & 4096       & 300   & -    & 69.0          \\
BYOL~\cite{byol_2020}          & ViT-S    & 1024       & 300   & 66.6 & 71.4          \\
SwAV~\cite{swav_2020}          & ViT-S    & 1024       & 300   & 64.7 & 71.8          \\
MoCo-v3~\cite{mocov3_2021}       & ViT-S    & 4096       & 300   & -    & 72.5          \\
DINO~\cite{dino_2021}          & ViT-S    & 256        & 100   & 69.9 & 73.8          \\
DINO~\cite{dino_2021}          & ViT-S    & 256        & 200   & 72.8 & 75.9          \\
\rowcolor{Light} \textbf{\method} (R50-ViT-S)  & ViT-S    & 256        & 100   & 70.8 & 74.4          \\
\rowcolor{Light} \textbf{\method} (R50-ViT-S)  & ViT-S    & 256        & 200   & \textbf{73.1} & \textbf{76.3} \\ \hline
MoCo-v3~\cite{mocov3_2021}       & ViT-B    & 4096       & 300   & -    & 76.5          \\
DINO~\cite{dino_2021}          & ViT-B    & 256        & 100   & 73.6 & 77.0          \\
DINO~\cite{dino_2021}          & ViT-B    & 256        & 200   & 75.1 & 77.7          \\
\rowcolor{Light} \textbf{\method} (R50-ViT-B)  & ViT-B    & 256        & 100   & 75.2 & 78.0 \\
\rowcolor{Light} \textbf{\method} (R50-ViT-B)  & ViT-B    & 256        & 200   & \textbf{76.0}    & \textbf{78.4} \\
\bottomrule
\end{tabular}}
\caption{Comparison of MOKD and SSL methods on ImageNet using $k$-NN and linear probing (LP) accuracy. Bold font indicates the best results. BS denotes batch size.}
\label{tab_compared_ssl}
\end{table}

\begin{table}[tp]
\centering
\setlength{\abovecaptionskip}{0.2cm}
\setlength{\tabcolsep}{0.4mm}{
\begin{tabular}{l|cc|cc|cc|cc}
\toprule
Method       & R50  & R34  & R50  & R18  & R101 & R34  & R101 & R18  \\ \hline
Supervised   & 76.2 & 75.0 & 76.2 & 72.1 & 77.0 & 75.0 & 77.0 & 72.1 \\
SEED~\cite{seed_2021}         & 67.4 & 58.5 & 67.4 & 57.6 & 70.3 & 61.6 & 70.3 & 58.9 \\
ReKD~\cite{rekd_2022}         & - & - & 67.6 & 59.6 & - & - & 69.7 & 59.7 \\
MCL~\cite{mcl_2022}          & 61.8 & 55.0 & 59.5 & 51.4 & 62.8 & 55.6 & 60.8 & 51.8 \\
DisCo~\cite{disco_2022}        & 67.4 & 62.5 & 67.4 & 60.6 & 69.1 & 64.4 & 69.1 & 62.3 \\
\rowcolor{Light} \textbf{\method} & 72.3 & \textbf{67.0} & 72.0 & \textbf{63.4} & 74.7 & \textbf{67.6} & 74.5 & \textbf{63.6} \\
\bottomrule
\end{tabular}}
\caption{Comparison of MOKD and SSL-KD methods on ImageNet using the linear probing accuracy.}
\label{tab_compared_sslkd}
\end{table}

\noindent
\textbf{Compared with SSL-KD Methods.}
\method is also compared with other SSL-KD methods, including SEED\cite{seed_2021}, ReKD\cite{rekd_2022}, MCL\cite{mcl_2022}, and DisCo\cite{disco_2022}.
Following SEED and DisCo, R101 and R50 are used as the larger model (or teacher model), and R34 and R18 are employed as the smaller model (or student model).
The linear probing accuracy is reported.
The results are shown in \cref{tab_compared_sslkd}.
\method achieves the best performance for all models, outperforming the state-of-the-art method DisCo.
For example, with R50-R34, \method achieves 67.0\% for R34, which outperforms DisCo by a margin of 4.5\%.

\subsection{Semi-Supervised Learning}
\label{subsec_exp_semisupervised}

In this part, we evaluate the performance of \method under the semi-supervised setting.
Specifically, we use the 1\% and 10\% subsets \cite{simclrv2_2020} of the ImageNet \cite{imagenet_2015} training set for fine-tuning, which follows the semi-supervised protocol in \cite{simclr_2020}.
Models are fine-tuned with 1024 batch size for 60 epochs and 30 epochs on 1\% and 10\% subsets, respectively.
The top-1 accuracy is employed.
The results are reported in \cref{tab_semi}.
Fine-tuning using 1\% and 10\% training data, \method improves the performance of the two models with a large margin compared with the models pre-trained independently.

\begin{table}[t]
\centering
\setlength{\abovecaptionskip}{0.2cm}
\setlength{\tabcolsep}{0.3mm}{
\begin{tabular}{ll|cc|>{\columncolor{Light}}c>{\columncolor{Light}}c|cc|>{\columncolor{Light}}c>{\columncolor{Light}}c}
\toprule
\multicolumn{2}{c}{}\vline & \multicolumn{4}{c}{1\%}\vline & \multicolumn{4}{c}{10\%} \\ \hline
\multicolumn{2}{c}{Backbone} \vline & \multicolumn{2}{c}{Independent} \vline & \multicolumn{2}{c}{\method}\vline & \multicolumn{2}{c}{Independent}\vline & \multicolumn{2}{c}{\method} \\ \hline
Net1  & Net2  & Net1 & Net2 & Net1 & Net2 & Net1 & Net2 & Net1 & Net2 \\ \hline
R50   & ViT-S & 49.5 & 44.6 & \textbf{53.8} & \textbf{44.9} & 67.3 & 67.8 & \textbf{68.7} & \textbf{68.2} \\
R50   & ViT-B & 49.5 & 57.1 & \textbf{57.2} & \textbf{58.3} & 67.3 & 73.4 & \textbf{70.2} & \textbf{74.3} \\
R101  & ViT-S & 54.8 & 44.6 & \textbf{56.9} & \textbf{44.8} & 70.7 & 67.8 & \textbf{71.0} & \textbf{68.8} \\
R101  & ViT-B & 54.8 & 57.1 & \textbf{57.1} & \textbf{63.9} & 70.7 & 73.4 & \textbf{71.0} & \textbf{74.7} \\ \hline
R50   & R34   & 49.5 & 44.5 & \textbf{49.8} & \textbf{46.1} & 67.3 & 62.9 & 67.3 & \textbf{63.5} \\
R50   & R18   & 49.5 & 35.8 & \textbf{50.0} & \textbf{40.8} & 67.3 & 56.0 & 67.3 & \textbf{56.7} \\
R101  & R34   & 54.8 & 44.5 & 54.8 & \textbf{47.0} & 70.7 & 62.9 & \textbf{70.8} & \textbf{63.7} \\
R101  & R18   & 54.8 & 35.8 & \textbf{54.9} & \textbf{41.4} & 70.7 & 56.0 & 70.7 & \textbf{56.8} \\
R50   & R50   & 49.5 & 49.5 & \textbf{50.1} & \textbf{49.8} & 67.3 & 67.3 & \textbf{67.6} & \textbf{67.4} \\
R18   & R18   & 35.8 & 35.8 & \textbf{36.5} & \textbf{36.4} & 56.0 & 56.0 & \textbf{56.3} & \textbf{56.3} \\ \hline
ViT-S & ViT-T & 44.6 & 19.2 & \textbf{44.9} & \textbf{20.0} & 67.8 & 55.2 & \textbf{68.2} & \textbf{55.8} \\
ViT-S & ViT-S & 44.6 & 44.6 & \textbf{44.8} & \textbf{44.7} & 67.8 & 67.8 & \textbf{68.3} & \textbf{68.3} \\
\bottomrule
\end{tabular}}
\caption{Semi-supervised learning on ImageNet subset. The top-1 accuracy is reported.}
\label{tab_semi}
\end{table}

\subsection{Transfer to Cifar10/Cifar100}
\label{subsec_exp_cifar}

We further fine-tune the pre-trained models on Cifar10 and Cifar100 \cite{cifar10} datasets to analyze the generalization of representations obtained by \method.
The models are fine-tuned with 1024 batch size for 100 epochs.
The top-1 accuracy is employed.
As shown in \cref{tab_cifar}, \method surpasses the independently pre-training baseline with different models on both Cifar10 and Cifar100.
This experiment shows the good generalization ability of \method.

\begin{table}[t]
\centering
\setlength{\abovecaptionskip}{0.2cm}
\setlength{\tabcolsep}{0.3mm}{
\begin{tabular}{ll|cc|>{\columncolor{Light}}c>{\columncolor{Light}}c|cc|>{\columncolor{Light}}c>{\columncolor{Light}}c}
\toprule
\multicolumn{2}{c}{}\vline & \multicolumn{4}{c}{Cifar10}\vline & \multicolumn{4}{c}{Cifar100} \\ \hline
\multicolumn{2}{c}{Backbone} \vline & \multicolumn{2}{c}{Independent} \vline & \multicolumn{2}{c}{\method}\vline & \multicolumn{2}{c}{Independent}\vline & \multicolumn{2}{c}{\method} \\ \hline
Net1  & Net2  & Net1 & Net2 & Net1 & Net2 & Net1 & Net2 & Net1 & Net2 \\ \hline
R50   & ViT-S & 97.3 & 98.6 & \textbf{97.6} & \textbf{98.7} & 85.3 & 88.8 & \textbf{86.2} & \textbf{88.9} \\
R50   & ViT-B & 97.3 & 98.8 & \textbf{97.6} & \textbf{99.2} & 85.3 & 91.0 & \textbf{85.6} & \textbf{91.3} \\
R101  & ViT-S & 98.2 & 98.6 & \textbf{98.4} & \textbf{98.8} & 87.6 & 88.8 & \textbf{87.8} & \textbf{89.0} \\
R101  & ViT-B & 98.2 & 98.8 & \textbf{98.3} & \textbf{99.0} & 87.6 & 91.0 & \textbf{87.8} & \textbf{91.4} \\ \hline
R50   & R34   & 97.3 & 97.0 & \textbf{97.5} & \textbf{97.2} & 85.3 & 83.3 & 85.0 & \textbf{83.7} \\
R50   & R18   & 97.3 & 95.6 & \textbf{97.5} & \textbf{96.1} & 85.3 & 80.2 & \textbf{85.2} & \textbf{81.1} \\
R101  & R34   & 98.2 & 97.0 & \textbf{98.3} & \textbf{97.3} & 87.6 & 83.3 & 87.6 & \textbf{83.9} \\
R101  & R18   & 98.2 & 95.6 & 98.1 & \textbf{96.0} & 87.6 & 80.2 & 87.3 & \textbf{80.8} \\
R50   & R50   & 97.3 & 97.3 & \textbf{97.7} & \textbf{97.6} & 85.3 & 85.3 & \textbf{85.6} & \textbf{85.4} \\
R18   & R18   & 95.6 & 95.6 & \textbf{95.9} & \textbf{95.8} & 80.2 & 80.2 & \textbf{80.6} & \textbf{80.6} \\ \hline
ViT-S & ViT-T & 98.6 & 97.2 & \textbf{98.7} & \textbf{97.4} & 88.8 & 84.1 & \textbf{89.0} & \textbf{84.4} \\
ViT-S & ViT-S & 98.6 & 98.6 & \textbf{98.7} & \textbf{98.7} & 88.8 & 88.8 & \textbf{89.0} & \textbf{88.9} \\
\bottomrule
\end{tabular}}
\caption{Tranfer learning on Cifar10 and Cifar100. The top-1 accuracy is reported.}
\label{tab_cifar}
\end{table}

\subsection{Transfer to Detection and Segmentation}
\label{subsec_exp_detection}

In this part, we evaluate the representation of \method on dense prediction tasks, i.e., object detection and instance segmentation, on MS COCO \cite{coco_2014} datasets.
We use the train2017 set for training and evaluate on the val2017 set.
Following \cite{seed_2021}, C4-based Mask R-CNN \cite{maskrcnn_2017} is used for objection detection and instance segmentation on COCO.
And R34 is used as backbone, which is initialized by the pre-trained models.
Our implementation is based on detectron2~\cite{wu2019detectron2}.
The experimental results are shown in \cref{tab_det}.
It can be seen that \method achieves the best performance, which demonstrates that \method has good generalization ability on dense prediction tasks.

\begin{table}[t]
\centering
\setlength{\abovecaptionskip}{0.2cm}
\setlength{\tabcolsep}{0.35mm}{
\begin{tabular}{l|cc|ccc|ccc}
\toprule
\multirow{2}{*}{Method} & \multirow{2}{*}{Net1} & \multirow{2}{*}{Net2} & \multicolumn{3}{c}{Detection} \vline & \multicolumn{3}{c}{Segmentation} \\
 &  &  & $\rm{AP}^{\rm{b}}$ & $\rm{AP}_{50}^{\rm{b}}$ & $\rm{AP}_{75}^{\rm{b}}$ & $\rm{AP}^{\rm{s}}$ & $\rm{AP}_{50}^{\rm{s}}$ & $\rm{AP}_{75}^{\rm{s}}$ \\ \hline
MoCov2~\cite{mocov2_2020} & - & R34 & 38.1 & 56.8 & 40.7 & 33.0 & 53.2 & 35.3 \\
DINO~\cite{dino_2021} & - & R34 & 39.6 & 58.6 & 42.5 & 34.2 & 55.3 & 36.4 \\ \hline
SEED~\cite{seed_2021} & R50 & R34 & 38.4 & 57.0 & 41.0 & 33.3 & 53.2 & 35.3 \\
MCL~\cite{mcl_2022} & R50 & R34 & 39.5 & 58.4 & 42.5 & 34.1 & 55.3 & 36.4 \\
DisCo~\cite{disco_2022} & R50 & R34 & 40.0 & 59.1 & 43.4 & 34.9 & 56.3 & 37.1 \\
\rowcolor{Light} \textbf{\method} & R50 & R34 & \textbf{40.3} & \textbf{59.7} & \textbf{43.7} & \textbf{34.9} & \textbf{56.5} & \textbf{37.1} \\  \hline
SEED~\cite{seed_2021} & R101 & R34 & 38.5 & 57.3 & 41.4 & 33.6 & 54.1 & 35.6 \\
MCL~\cite{mcl_2022} & R101 & R34 & 39.2 & 58.2 & 42.1 & 33.8 & 54.9 & 35.7 \\
DisCo~\cite{disco_2022} & R101 & R34 & 40.0 & 59.1 & 43.2 & 34.7 & 55.9 & 37.4 \\
\rowcolor{Light} \textbf{\method} & R101 & R34 & \textbf{40.3} & \textbf{59.3} & \textbf{43.5} & \textbf{34.8} & \textbf{56.2} & 36.9 \\
\bottomrule
\end{tabular}}
\caption{Object detection and instance segmentation on COCO with R34 as backbone. C4-based Mask R-CNN \cite{maskrcnn_2017} is adopted as the detector.}
\label{tab_det}
\end{table}

\subsection{Ablation Study}
\label{subsec_exp_ablation}

In this section, we analyze the influence of each component in \method.
The ImageNet100 dataset, which contains 100 randomly selected categories from ImageNet, is adopted to speed up the training time.
All models are trained on the ImageNet100 training set with 256 batch size and 200 epochs and tested on the validation set.
The linear probing top-1 accuracy is employed as the evaluation metric.

\noindent
\textbf{Effectiveness of Each Loss Term.}
There are four loss terms for each model in \method, i.e., $\mathcal{L}_{sm}$, $\mathcal{L}_{st}$, $\mathcal{L}_{cm}$, and $\mathcal{L}_{ct}$ in \cref{equ:l_sm}, \cref{equ:l_st}, \cref{equ:l_cm}, and \cref{equ:loss_ct}, respectively.
We use R50-ViT-S to analyze the influence of each loss term.
The results are shown in \cref{tab_ablation_loss}.
Note that the result with only $\mathcal{L}_{sm}$ is the DINO \cite{dino_2021} baseline that trains the two models independently.
With the addition of $\mathcal{L}_{cm}$, $\mathcal{L}_{st}$, and $\mathcal{L}_{ct}$, the performance improves gradually, which indicates that self-distillation and cross-distillation of MLP-head and T-Head can benefit the representation performance of \method.

\begin{table}[t]
\centering
\setlength{\abovecaptionskip}{0.2cm}
\begin{tabular}{cccc|cc}
\toprule
$\mathcal{L}_{sm}$ & $\mathcal{L}_{cm}$ & $\mathcal{L}_{st}$ & $\mathcal{L}_{ct}$ & R50 & ViT-S \\ \hline
$\checkmark$ & &  &  & 87.0 & 80.3 \\
$\checkmark$ & $\checkmark$ &  &  & 87.2 & 82.6 \\
$\checkmark$ & $\checkmark$ & $\checkmark$ &  & 87.7 & 83.5 \\
$\checkmark$ & $\checkmark$ & $\checkmark$ & $\checkmark$ & 88.3 & 84.6 \\
\bottomrule
\end{tabular}
\caption{Influence of each loss term.}
\label{tab_ablation_loss}
\end{table}

\noindent
\textbf{Influence of $\lambda _1$ and $\lambda _2$.}
$\lambda _1$ and $\lambda _2$ in \cref{equ:l_all} are the weights of cross-distillation loss of model1 and model2, respectively.
\cref{tab_ablation_lamda} shows the results.
As shown in the first row, when $\lambda _2\!=\!1$ for ViT-S, the performance of R50 get worse with the increase of $\lambda _1$, which indicates that it is better to set a small value for the model with better performance.
When $\lambda _1\!=\!0$, R50 is trained without cross-distillation and achieves insignificant improvement, which indicates that cross-distillation is essential for the model with better performance.
As shown in the third row, when $\lambda _1\!=\!0.1$ for R50, the performance of ViT-S improves with the increase of $\lambda _2$, which demonstrates that paying more emphasis on cross-distillation is beneficial to the model with inferior performance.
In this study, $\lambda _1$ and $\lambda _2$ are set to 1 and 0.1 for the larger and smaller models, respectively.
While for model pairs with the same backbone, $\lambda _1$ and $\lambda _2$ are both set to 1.

\begin{table}[t]
\centering
\setlength{\abovecaptionskip}{0.2cm}
\setlength{\tabcolsep}{0.6mm}{
\begin{tabular}{c|c|c|c|c}
\toprule
$\lambda _1$, $\lambda _2$ & 0, 1 & 0.1, 1 & 0.5, 1 & 1, 1 \\ \hline
R50, ViT-S & 87.2, 83.1 & \textbf{88.3}, \textbf{84.6} & 87.9, 84.5 & 87.4, 84.1 \\\hline\hline
$\lambda _1$, $\lambda _2$ & 0.1, 0 & 0.1, 0.1 & 0.1, 0.5 & 0.1, 1 \\\hline
R50, ViT-S & 87.4, 80.4 & 87.5, 81.3 & 87.9, 83.1 & \textbf{88.3}, \textbf{84.6} \\
\bottomrule
\end{tabular}}
\caption{Influence of cross-distillation loss weights.}
\label{tab_ablation_lamda}
\end{table}

\begin{table}[t]
\centering
\setlength{\abovecaptionskip}{0.2cm}
\begin{tabular}{c|c|c|c}
\toprule
Methods & Independent & \method & Independent \\ \hline
$k$-NN & 62.8, 69.9 & 67.1, 70.8 & 67.5, 70.8 \\ \hline
Consistency & 0.871 & 0.901 & 0.900 \\
\bottomrule
\end{tabular}
\caption{Prediction consistency between two models.}
\label{tab_ablation_consistency}
\end{table}

\subsection{Visualization and Analysis}
\label{subsec_exp_vis}

\begin{figure}
  \centering
  \begin{subfigure}{0.48\linewidth}
    \includegraphics[width=1.6in]{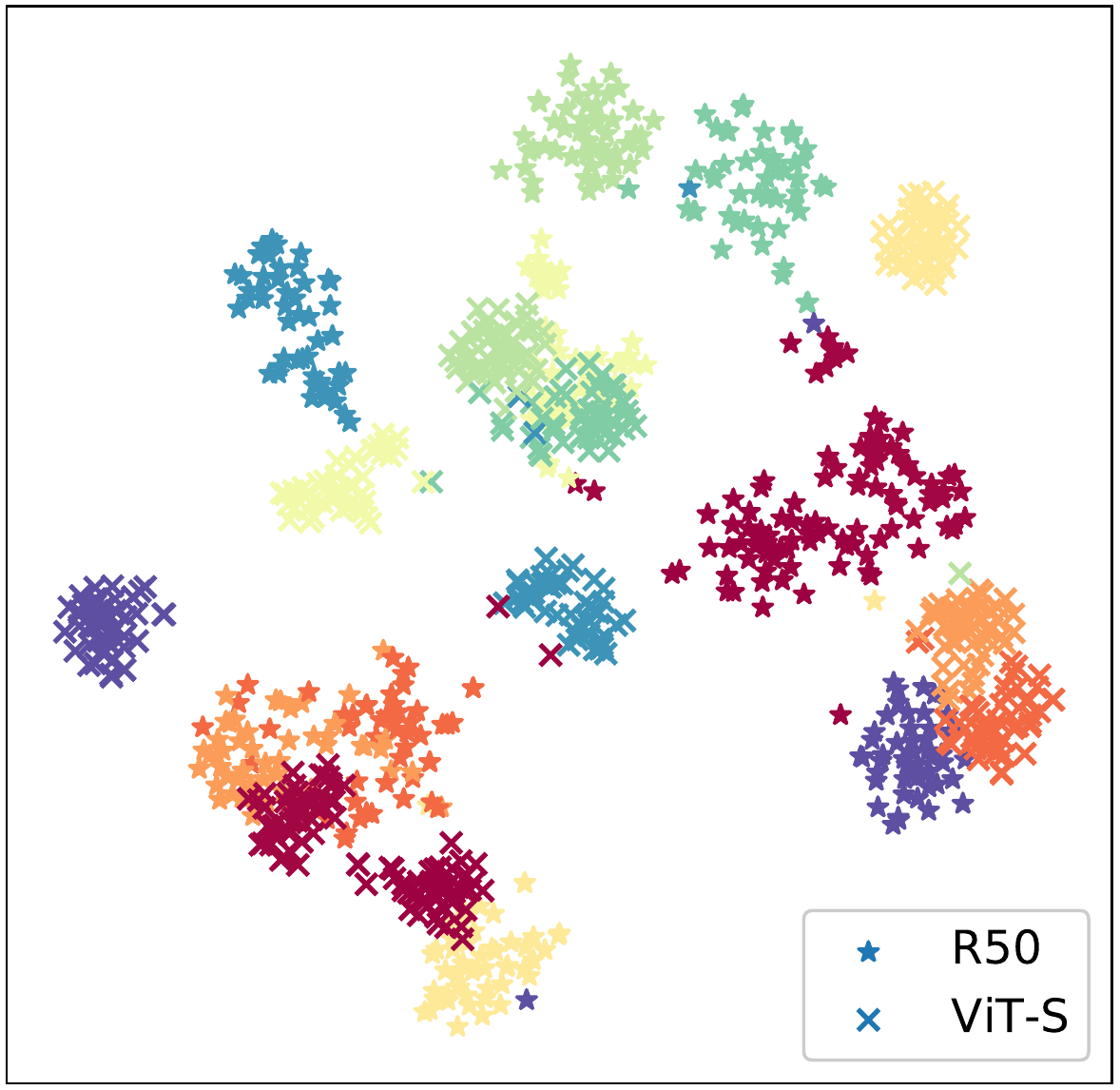}
    \caption{Independent (R50-ViT-S).}
    \label{fig_exp_tsne_a}
  \end{subfigure}
  \hfill
  \begin{subfigure}{0.48\linewidth}
    \includegraphics[width=1.6in]{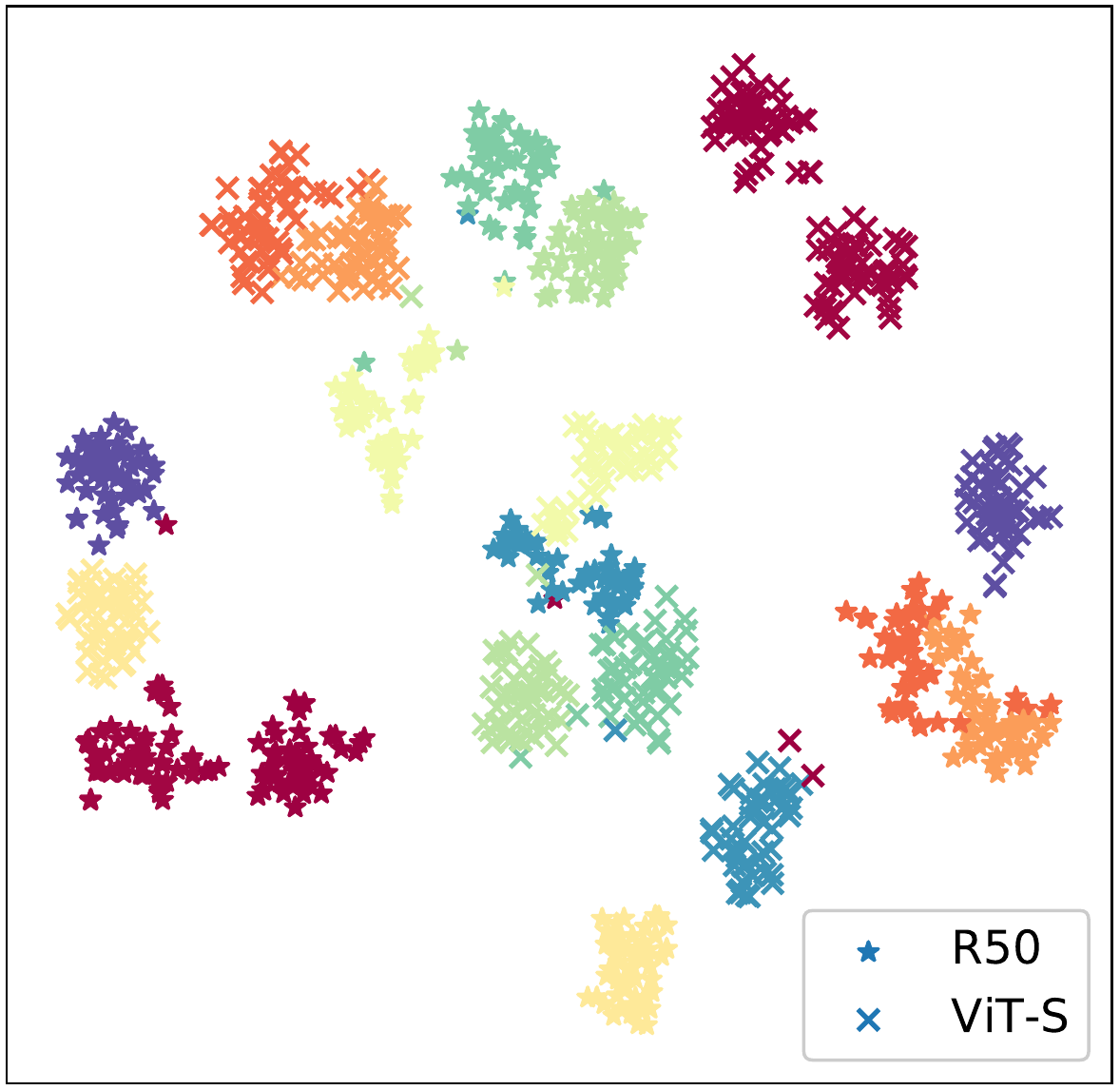}
    \caption{\method (R50-ViT-S).}
    \label{fig_exp_tsne_b}
  \end{subfigure}
  \caption{T-SNE \cite{tsne_2018} visualization of feature distributions on ImageNet100. Ten categories (shown in different colors) are randomly selected for better visualization.}
  \label{fig_exp_tsne}
\end{figure}

\noindent
\textbf{Does \method Make Models More Similar?}
In this part, we analyze the representations learned by \method.
First, we analyze if representations of two models trained by \method tend to be more similar.
To this end, the fraction of samples that two model pairs make the same prediction is calculated.
The R50-ViT-S configuration, which is pre-trained on ImageNet, is employed, and $k$-NN is applied on ImageNet100.
As shown in \cref{tab_ablation_consistency}, the prediction consistency rate of \method increases compared to the result of independent training in the second column.
However, this increase is mainly caused by the performance improvement of the two models compared to the consistency rate in the last column, which is obtained by replacing the R50 model trained by \method with an R50 model with relative accuracy trained by DINO.
The results verify that there is no significant tendency for representations of two models trained by \method to become more similar.
\cref{fig_exp_tsne} visualizes the feature distributions of the two models trained by \method.
The two models show different feature distributions, which further verifies that \method does not make models more similar.
More results can be found in \cref{fig_app_tsne} in \supp.

\begin{figure}
  \centering
  \begin{subfigure}{0.48\linewidth}
    \includegraphics[width=1.6in]{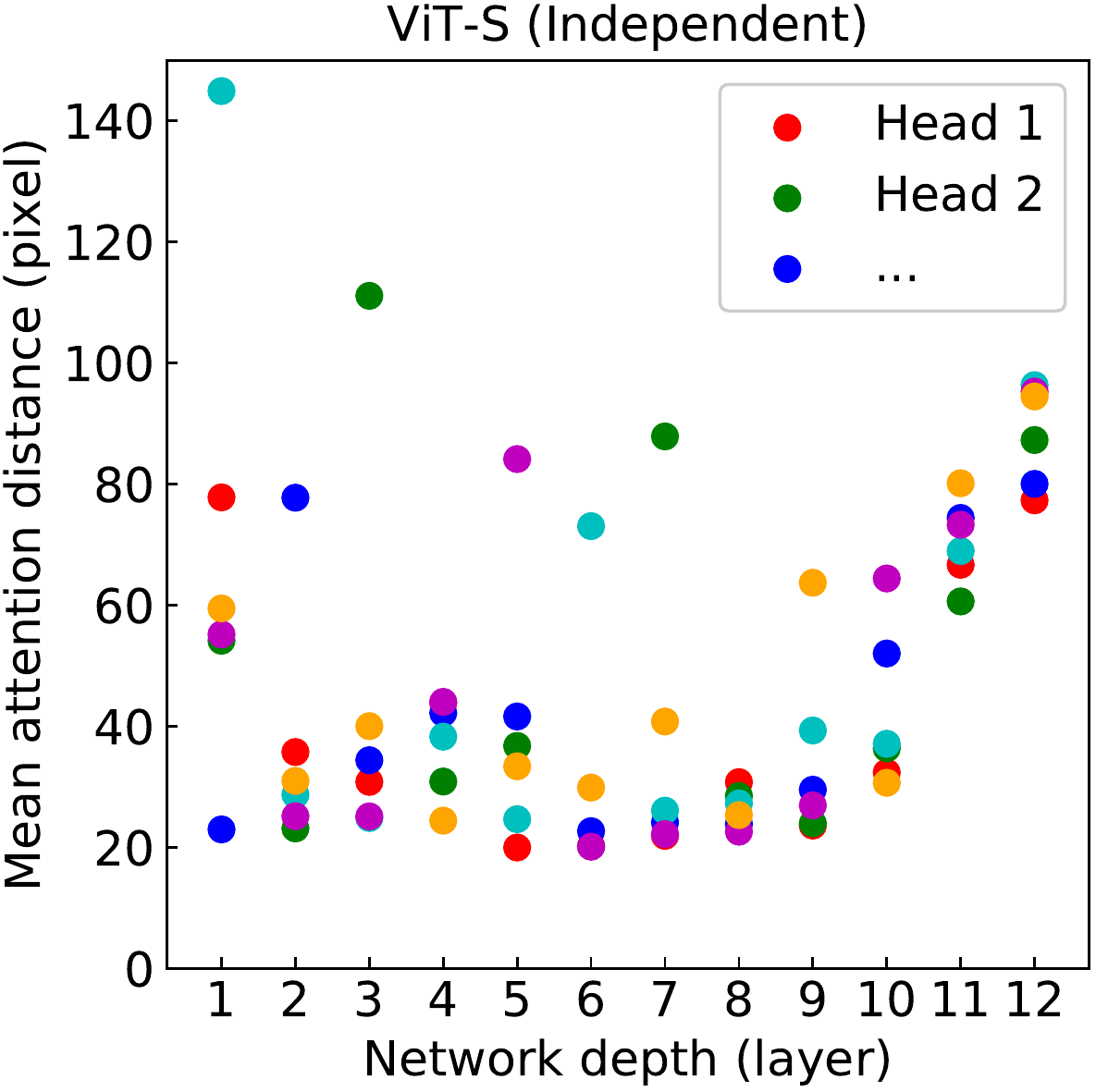}
    \caption{ViT-S trained independently.}
    \label{fig_exp_mad_a}
  \end{subfigure}
  \hfill
  \begin{subfigure}{0.48\linewidth}
    \includegraphics[width=1.6in]{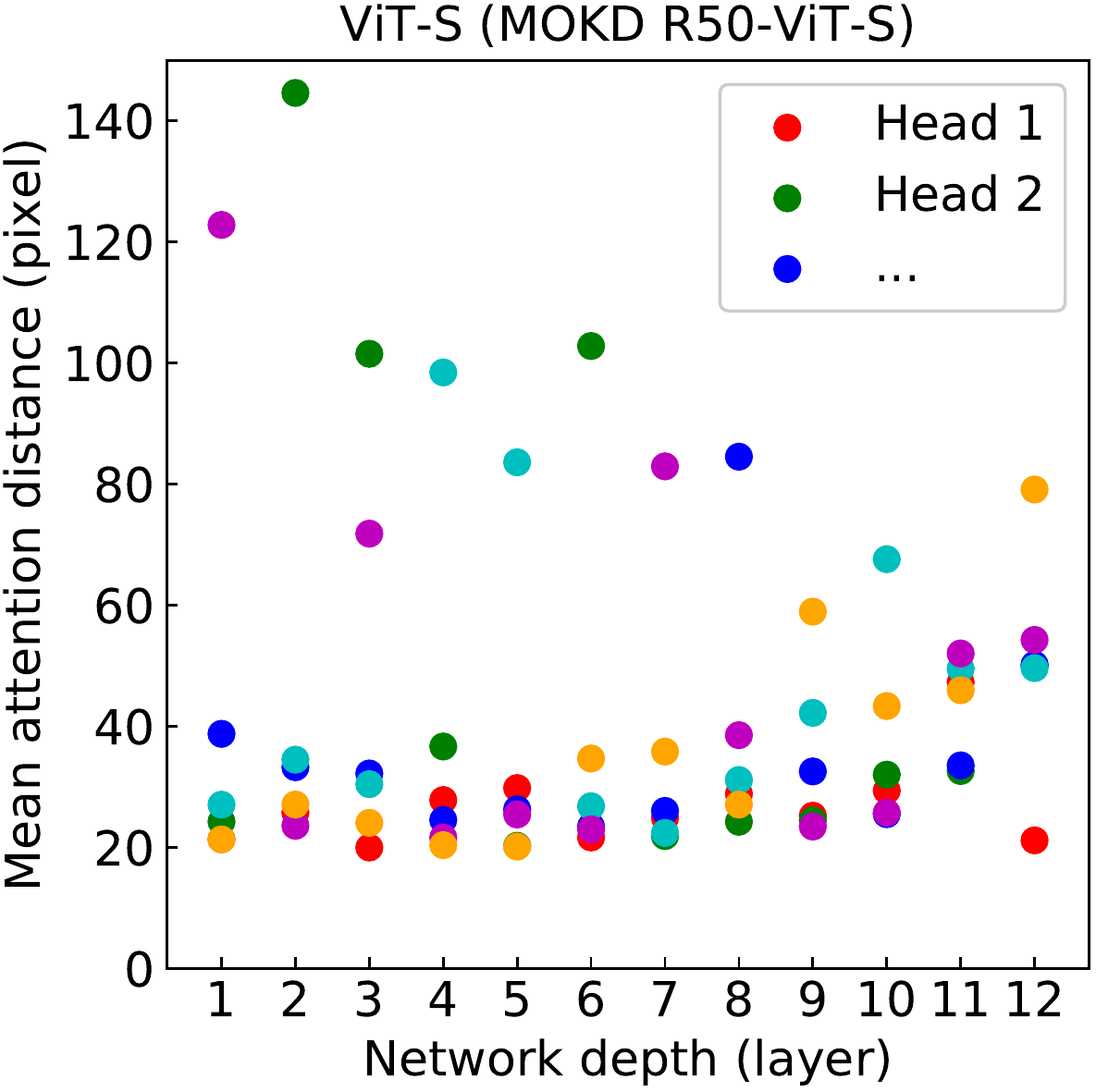}
    \caption{ViT-S trained by \method.}
    \label{fig_exp_mad_b}
  \end{subfigure}
  
  \begin{subfigure}{0.48\linewidth}
    \includegraphics[width=1.6in]{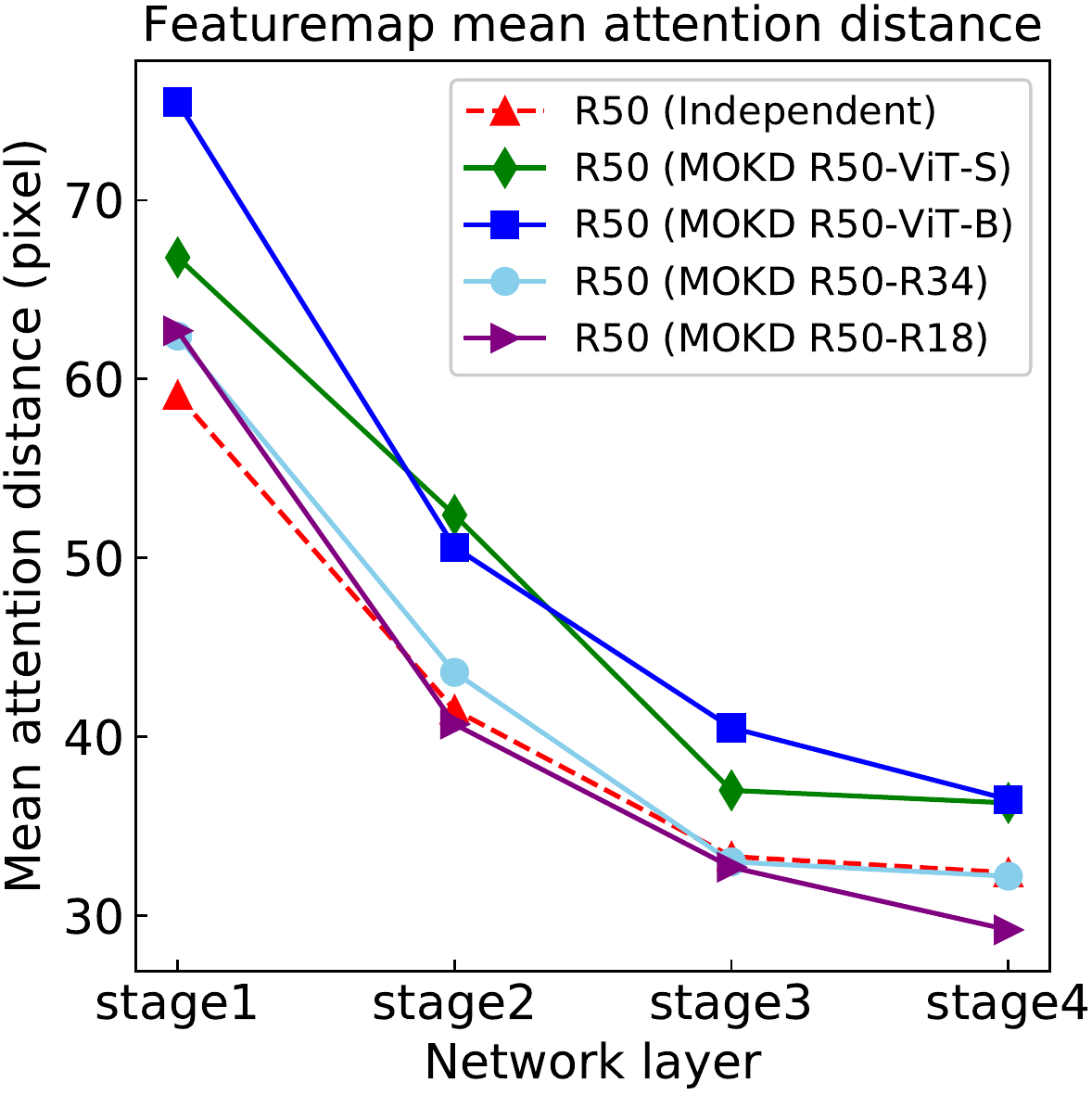}
    \caption{R50 models.}
    \label{fig_exp_mad_c}
  \end{subfigure}
  \hfill
  \begin{subfigure}{0.48\linewidth}
    \includegraphics[width=1.6in]{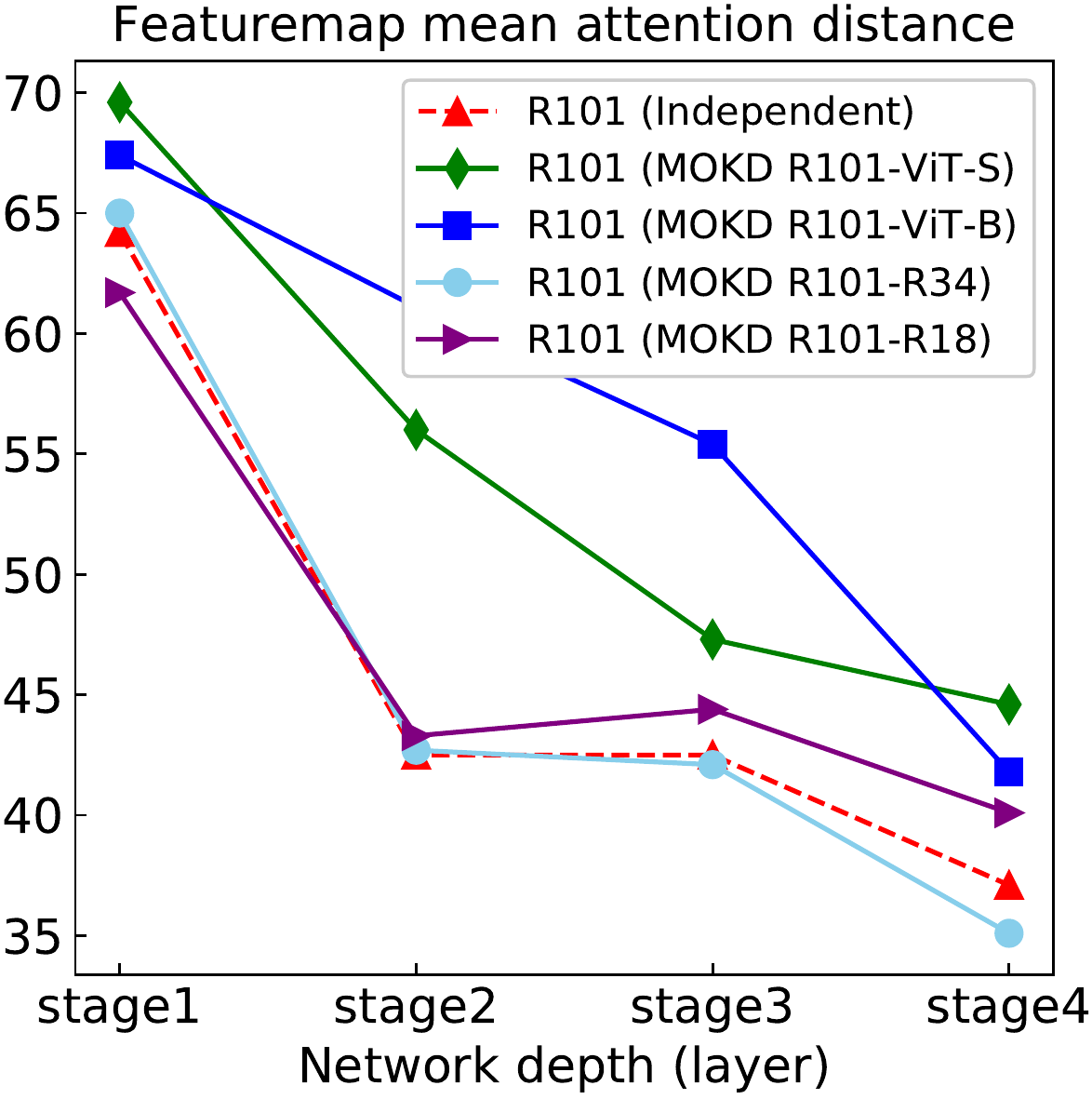}
    \caption{R101 models.}
    \label{fig_exp_mad_d}
  \end{subfigure}
  
  \caption{Mean attention distances \cite{vit_2021} of different models.}
  \label{fig_exp_mad}
\end{figure}

\noindent
\textbf{What Knowledge Is Learned in \method?}
We investigate the characteristic of features on each layer to analyze the difference between models trained independently and trained by \method.
Specifically, we calculate the mean attention distance (MAD) \cite{vit_2021} for ViT models.
For CNN models, a similar average distance can also be calculated on the feature map based on its self-attention weights.
As shown in \cref{fig_exp_mad}(a)(b), we found that the MAD on deep layers (layer 10-12) of the ViT-S model trained by \method (with R50) decreased compared to those of the ViT-S model trained independently, which indicates that the ViT-S model trained by \method turns to be more ``local" on deep layers.
The opposite phenomenon can be seen on ResNets.
As shown in \cref{fig_exp_mad}(c)(d), the MAD on each layer of ResNet models trained by \method (with ViTs) increase compared to those of ResNet models trained independently, which indicates that ResNet models trained by \method (with ViTs) turn to be more ``global".
However, this phenomenon is not shown in two ResNet models trained by \method.
That is, through \method, two heterogeneous models absorb knowledge from each other, i.e., ViT model learns more locality while CNN model learns more global information.
More results can be found in \cref{fig_app_mad} in \supp.

\section{Conclusions}
\label{sec:conclusion}

In this study, we propose the \method method, where two different models learn collaboratively through self-distillation and cross-distillation in a self-supervised manner.
Extensive experiments on different backbones and tasks demonstrate that \method can boost the feature representation performance of different models.
It achieves state-of-the-art performance for self-supervised knowledge distillation.
We hope this study could inspire boosting representation learning performance via knowledge interaction between heterogeneous models.

As an online knowledge distillation method, the main limitation of \method is that the larger model needs to be repeatedly trained for different smaller models, which requires more computation cost than offline knowledge distillation.
How to design an efficient \method can be further studied.
For example, introducing efficient fine-tuning methods \cite{adaptor_2019,vpt_2022} into \method may be a possible future study to realize efficient \method.

{\small
\bibliographystyle{ieee_fullname}
\bibliography{MOKD}
}

\clearpage
\renewcommand\thefigure{S\arabic{figure}}
\renewcommand\thetable{S\arabic{table}}  
\renewcommand\theequation{S\arabic{equation}}
\setcounter{equation}{0}
\setcounter{table}{0}
\setcounter{figure}{0}
\setcounter{section}{0}
\renewcommand\thesection{\Alph{section}}

\section*{Supplementary Material}

The implementation details are introduced first.
Then we show more visualization results for the feature distributions shown in Fig. \textcolor{red}{4} and mean attention distances shown in Fig. \textcolor{red}{5}.
Finally, we present more experimental results.

\section{Implementation Details}
\label{sec_sm_details}

\subsection{Pre-training on ImageNet}
\label{subsec:sm_pretrain}

\noindent
\textbf{Data Augmentation.}
The data augmentations consist of random cropping (with a scale of 0.25-1.0), resizing to $224 \times 224$, random horizontal flip, gaussian blur, and color jittering.
The local augmentations for the multi-crop strategy consist of random cropping (with a scale of 0.05-0.25), resizing to $96 \times 96$, random horizontal flip, gaussian blur, and color jittering.
Two global views and eight local views are used in all pre-taining experiments.

\noindent
\textbf{Training.}
During the pre-training procedure, we follow the most hyper-parameters setting of DINO \cite{dino_2021}.
Without a specific statement, the default batch size is 256.
The SGD and AdamW \cite{adamw_2018} optimizers are used for ResNet and ViT, respectively.
The learning rate is linearly warmed up to its base value during the first 10 epochs.
And the base learning rate is set to $0.1$ and $0.0003$ for ResNet and ViT, respectively.
After the warm-up procedure, the learning rate is decayed with a cosine schedule \cite{cosine_decay_2016}.
The weight decay is set to $1e-4$ and $0.04$ for ResNet and ViT, respectively.
For the temperatures, $\tau$ is set to $0.1$, and a linear warm-up from $0.04$ to $0.07$ is set to $\tau'$ during the first 30 epochs.
Following DINO \cite{dino_2021}, the centering operation is applied to the output of the momentum encoders to avoid collapse.
$\lambda _1$ and $\lambda _2$ in Eq. (\textcolor{red}{9}) are set to 1 and 0.1 for the larger and smaller models, respectively.
While for model pairs with the same backbone, $\lambda _1$ and $\lambda _2$ are set to 1.

\subsection{$k$-NN and Linear Probing on ImageNet}
\label{subsec:sm_knnlinear}

After pre-training, the $k$-NN and linear probing are employed to evaluate the representation performance.
For $k$-NN, it is implemented based on DINO \cite{dino_2021}.
We report the best result among $k=10,20,100,200$.

For linear probing, a linear classifier added to the frozen backbone is trained \cite{moco_2020}.
The linear classifier is trained with the SGD optimizer and a batch size of $2048$ for $100$ epochs on the ImageNet training set.
The learning rate is linearly warmed up to its base value during the first 10 epochs.
And the base learning rate is set to $0.08$ and $0.008$ for ResNet and ViT, respectively.
After the warm-up stage, the learning rate is decayed with a cosine schedule \cite{cosine_decay_2016}.
Weight decay is not used.
The input resolution is $224\times224$ during training and testing.

\subsection{Semi-Supervised Learning on ImageNet}
\label{subsec:sm_semisl}

We use the 1\% and 10\% subsets of the ImageNet \cite{imagenet_2015} training set for fine-tuning, which follows the semi-supervised protocol in \cite{simclr_2020}.
The same splits of 1\% and 10\% of ImageNet training set in \cite{simclrv2_2020} are used.
Models are fine-tuned with 1024 batch size for 60 epochs and 30 epochs on 1\% and 10\% subsets, respectively.
The SGD optimizer is adopted.
The learning rate is linearly warmed up to its base value during the first 5 epochs.
And the best base learning rate is searched for each model.
After the warm-up procedure, the learning rate is decayed with a cosine schedule \cite{cosine_decay_2016}.
The input resolution is $224\times224$ during training and testing.

\subsection{Fine-tuning on Cifar10/Cifar100}
\label{subsec:sm_cifar}

Most settings keep the same with the experiments of semi-supervised learning on ImageNet except for the training epochs.
On Cifar10/100, the warming-up epoch is set to 10, and the training epoch is set to 100.

\subsection{Fine-tuning on COCO}
\label{subsec:sm_coco}

Following \cite{seed_2021}, The C4-based Mask R-CNN \cite{maskrcnn_2017} detector is used for objection detection and instance segmentation on COCO.
The model is trained for 180k iterations, i.e., the 2$\times$ schedule.
The SGD optimizer is adopted.
The initial learning rate is set to 0.02.
The scale of images for training is set as [600, 800] and 800 at inference.
We report $\rm{AP}^{\rm{b}}$, $\rm{AP}_{50}^{\rm{b}}$, and $\rm{AP}_{75}^{\rm{b}}$ for object detection and $\rm{AP}^{\rm{s}}$, $\rm{AP}_{50}^{\rm{s}}$, and $\rm{AP}_{75}^{\rm{s}}$ for instance segmentation.

\section{More Visualizations}
\label{sec_sm_Visualizations}

\subsection{T-SNE Visualization of Feature Distribution}
\label{subsec:sm_tsne}

\begin{figure*}
  \centering
  \begin{subfigure}{0.23\linewidth}
    \includegraphics[width=1.6in]{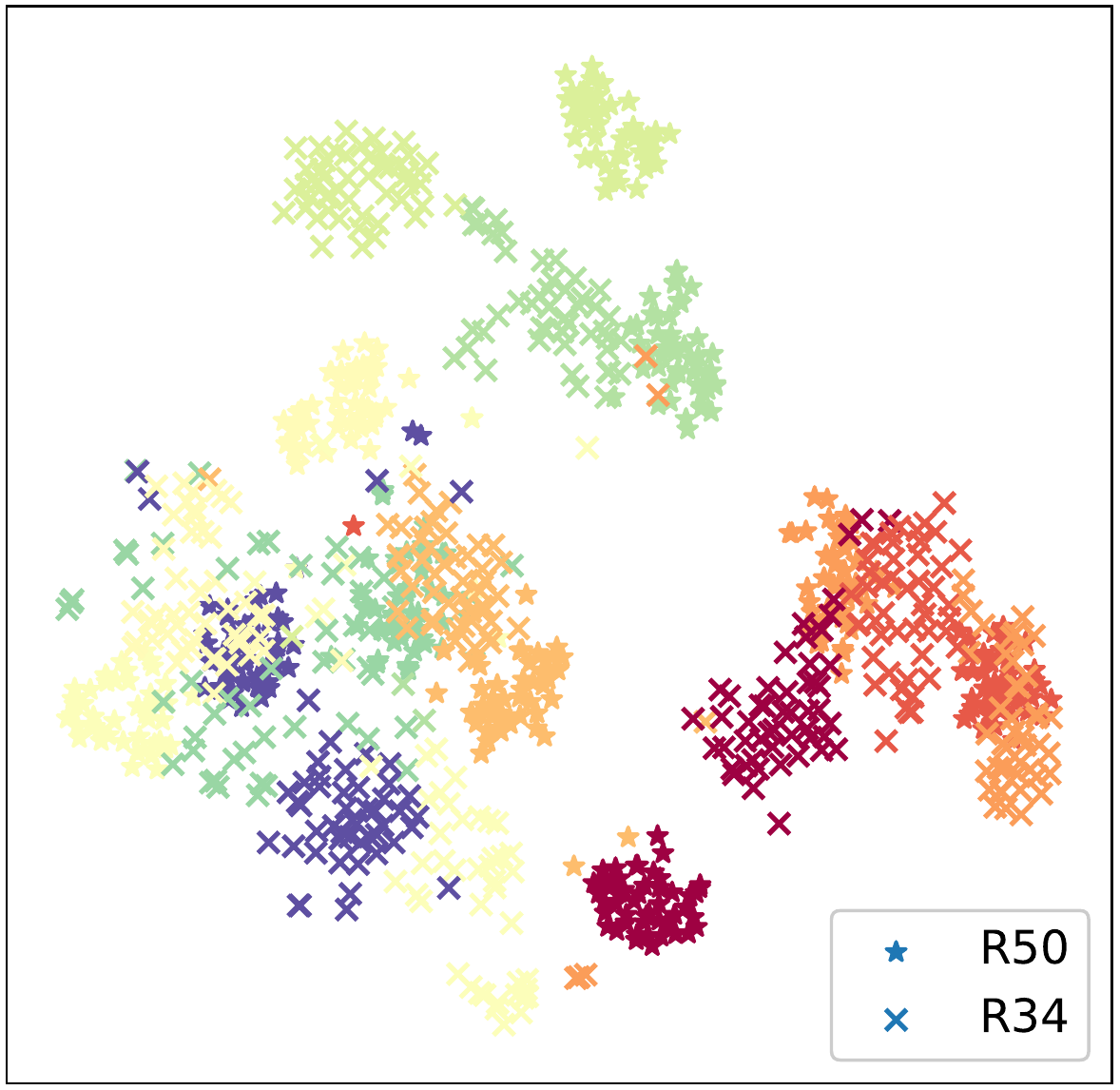}
    \caption{Independent (R50-R34).}
    \label{fig_app_tsne_a}
  \end{subfigure}
  \hfill
  \begin{subfigure}{0.23\linewidth}
    \includegraphics[width=1.6in]{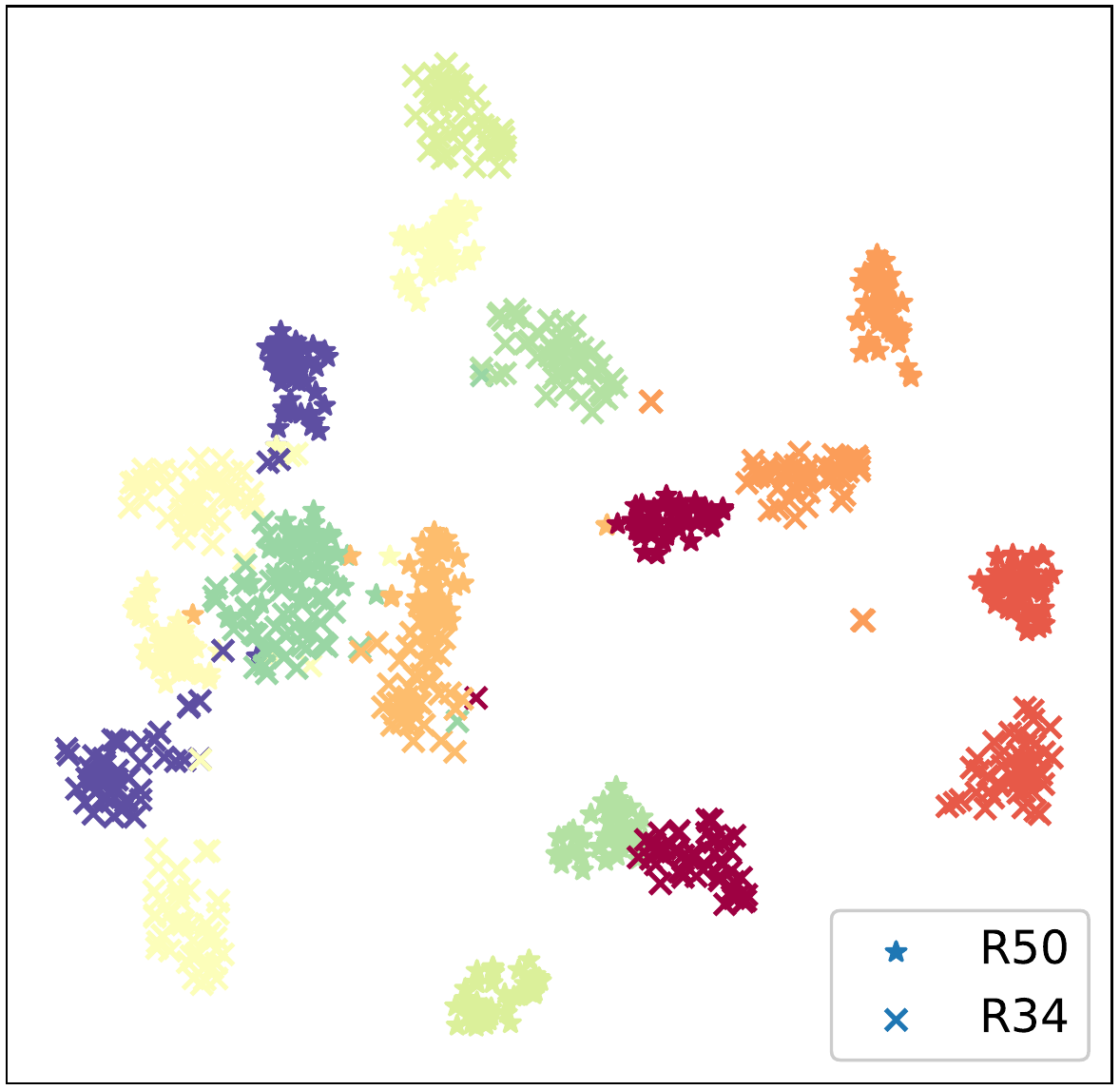}
    \caption{\method (R50-R34).}
    \label{fig_app_tsne_b}
  \end{subfigure}
  \hfill
  \begin{subfigure}{0.23\linewidth}
    \includegraphics[width=1.6in]{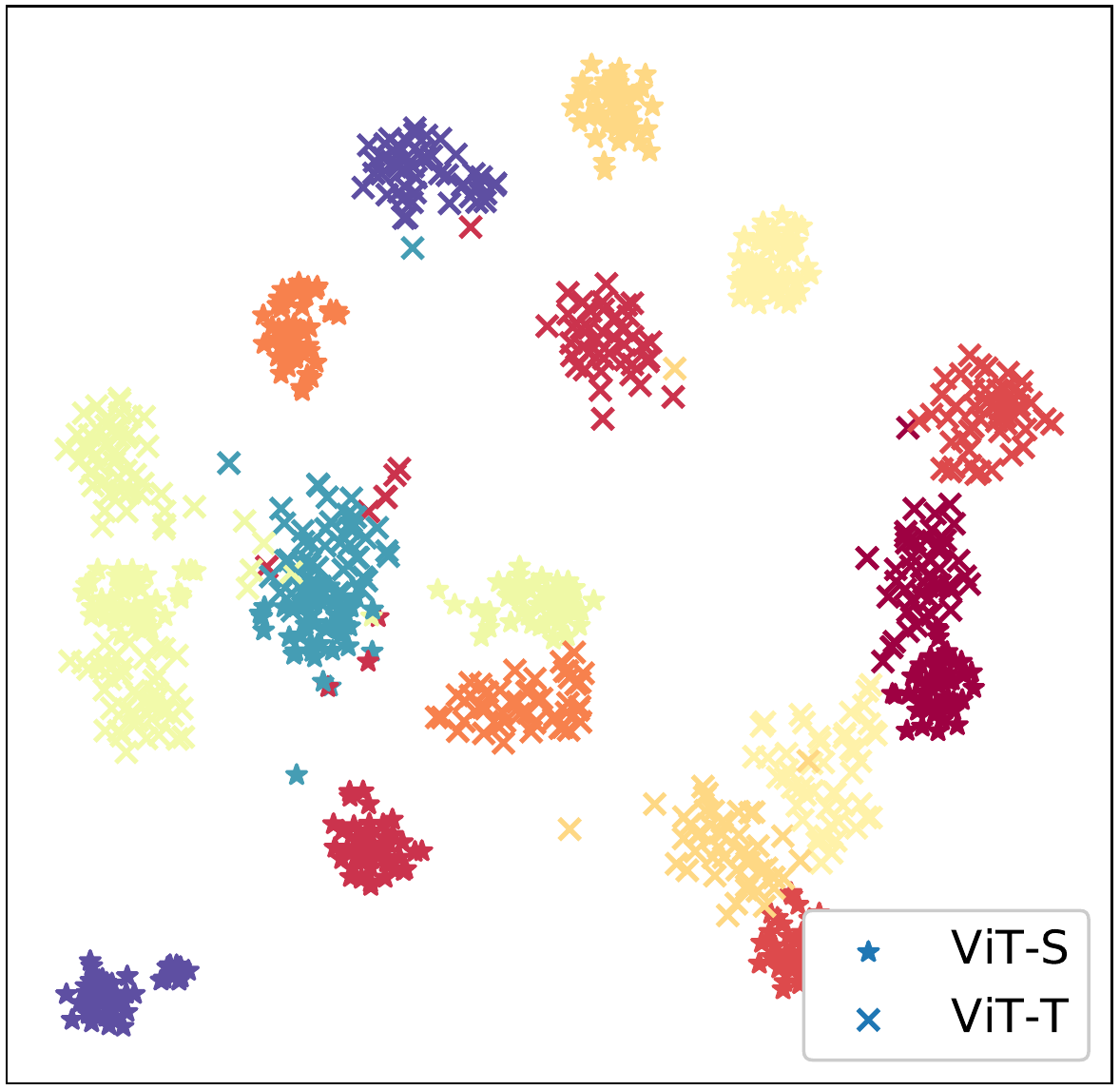}
    \caption{Independent (ViT-S-ViT-T).}
    \label{fig_app_tsne_c}
  \end{subfigure}
  \hfill
  \begin{subfigure}{0.23\linewidth}
    \includegraphics[width=1.6in]{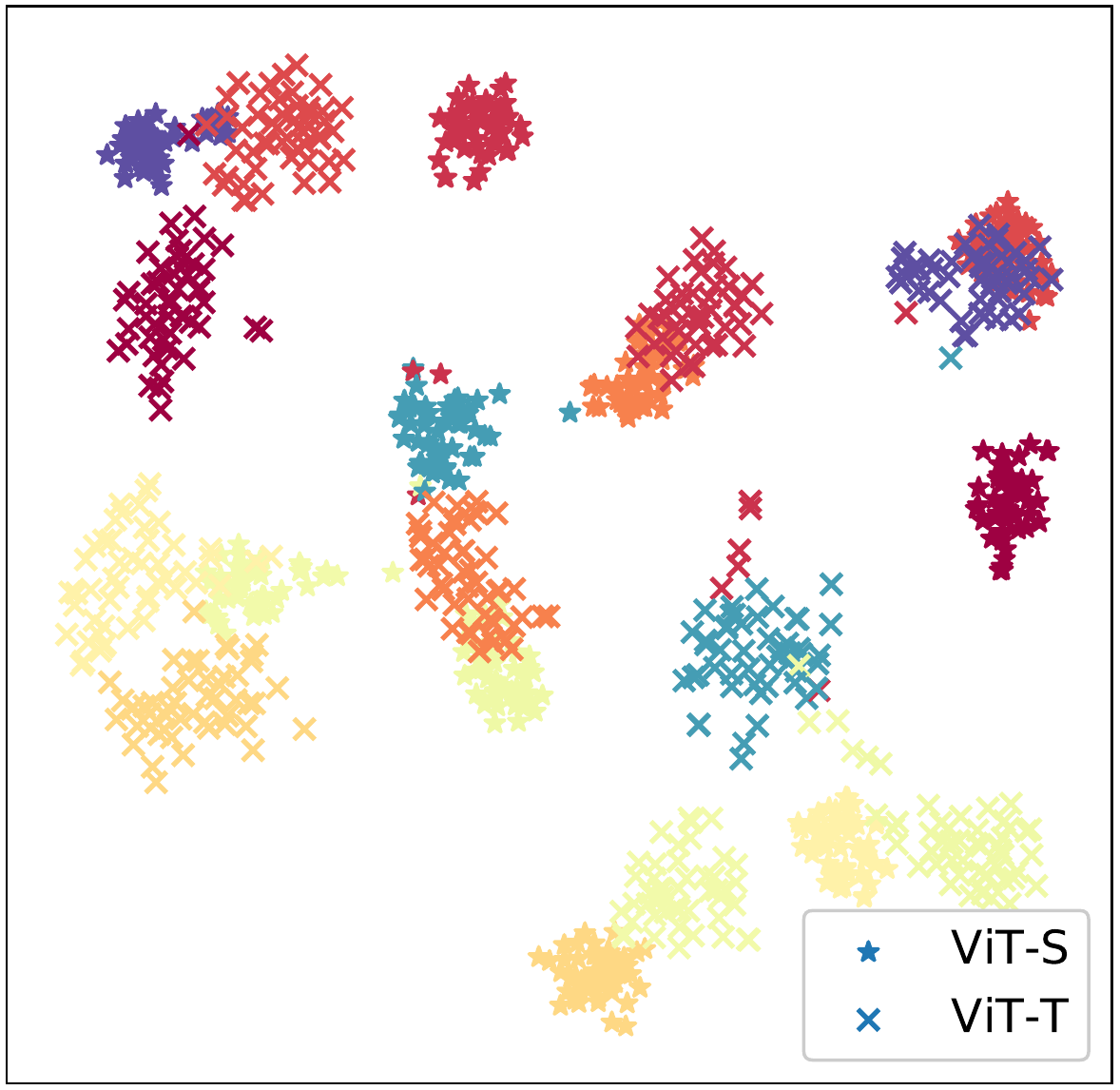}
    \caption{\method (ViT-S-ViT-T).}
    \label{fig_app_tsne_d}
  \end{subfigure}
  \caption{T-SNE \cite{tsne_2018} visualization of feature distributions on ImageNet100. Ten categories (shown in different colors) are randomly selected for better visualization. Different color denotes different category and different marker denotes different model.}
  \label{fig_app_tsne}
\end{figure*}

Following Fig. \textcolor{red}{4}, more model pairs, including two ResNets and two ViTs are visualized.
As shown in \cref{fig_app_tsne}, the model pairs (R50-R34 and ViT-S-ViT-T) trained by \method also show different feature distributions, demonstrating that \method does not make models more similar.
In addition, we can see that the feature distributions of R34 and ViT-T get better when trained with \method.

\begin{figure*}
  \centering
  \begin{subfigure}{0.23\linewidth}
    \includegraphics[width=1.6in]{figures/fig_exp_mad_vits_independent.pdf}
    \caption{ViT-S (Independent).}
    \label{fig_app_mad_a}
  \end{subfigure}
  \hfill
  \begin{subfigure}{0.23\linewidth}
    \includegraphics[width=1.6in]{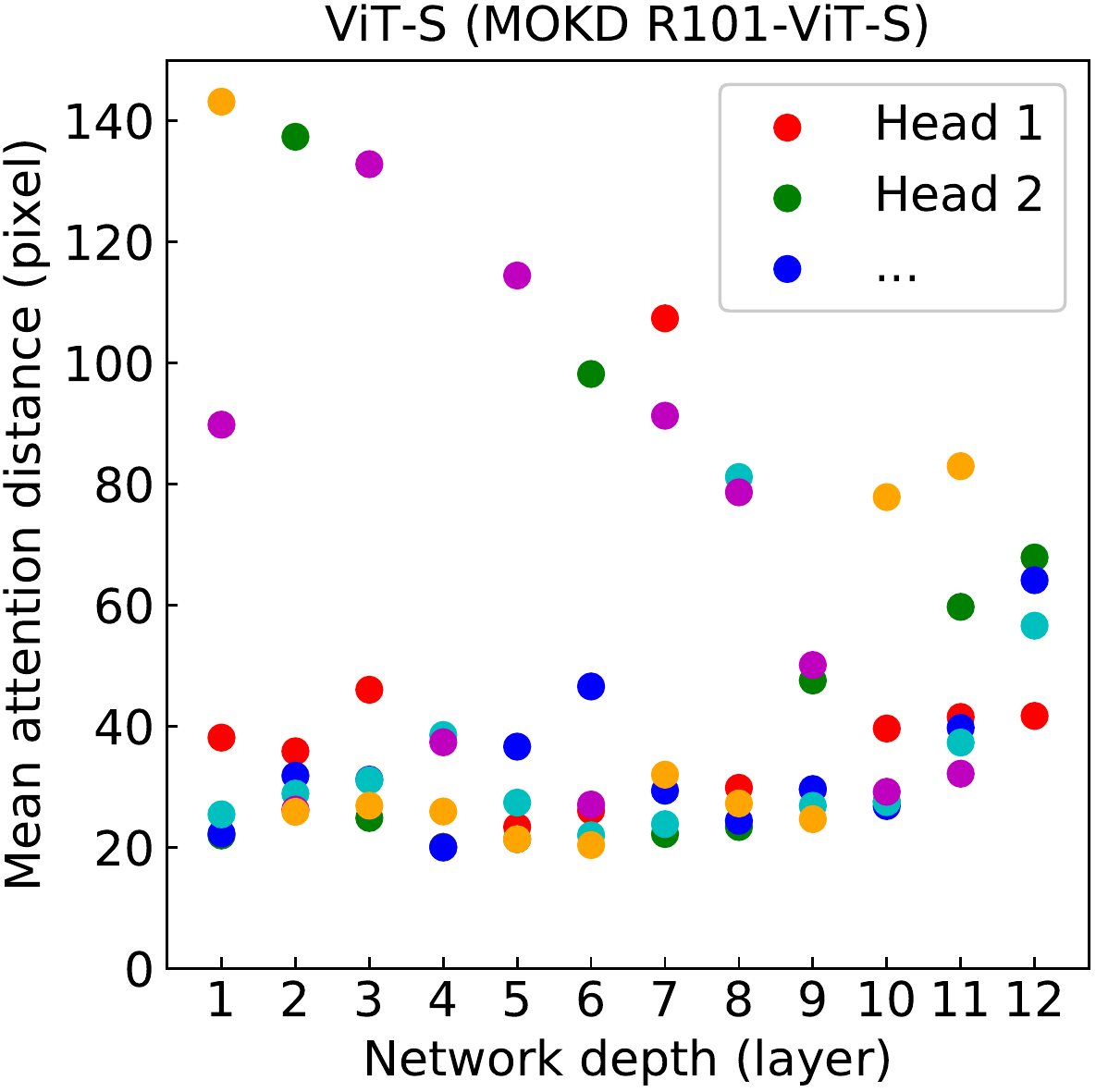}
    \caption{ViT-S (\method with R101).}
    \label{fig_app_mad_b}
  \end{subfigure}
  \hfill
  \begin{subfigure}{0.23\linewidth}
    \includegraphics[width=1.6in]{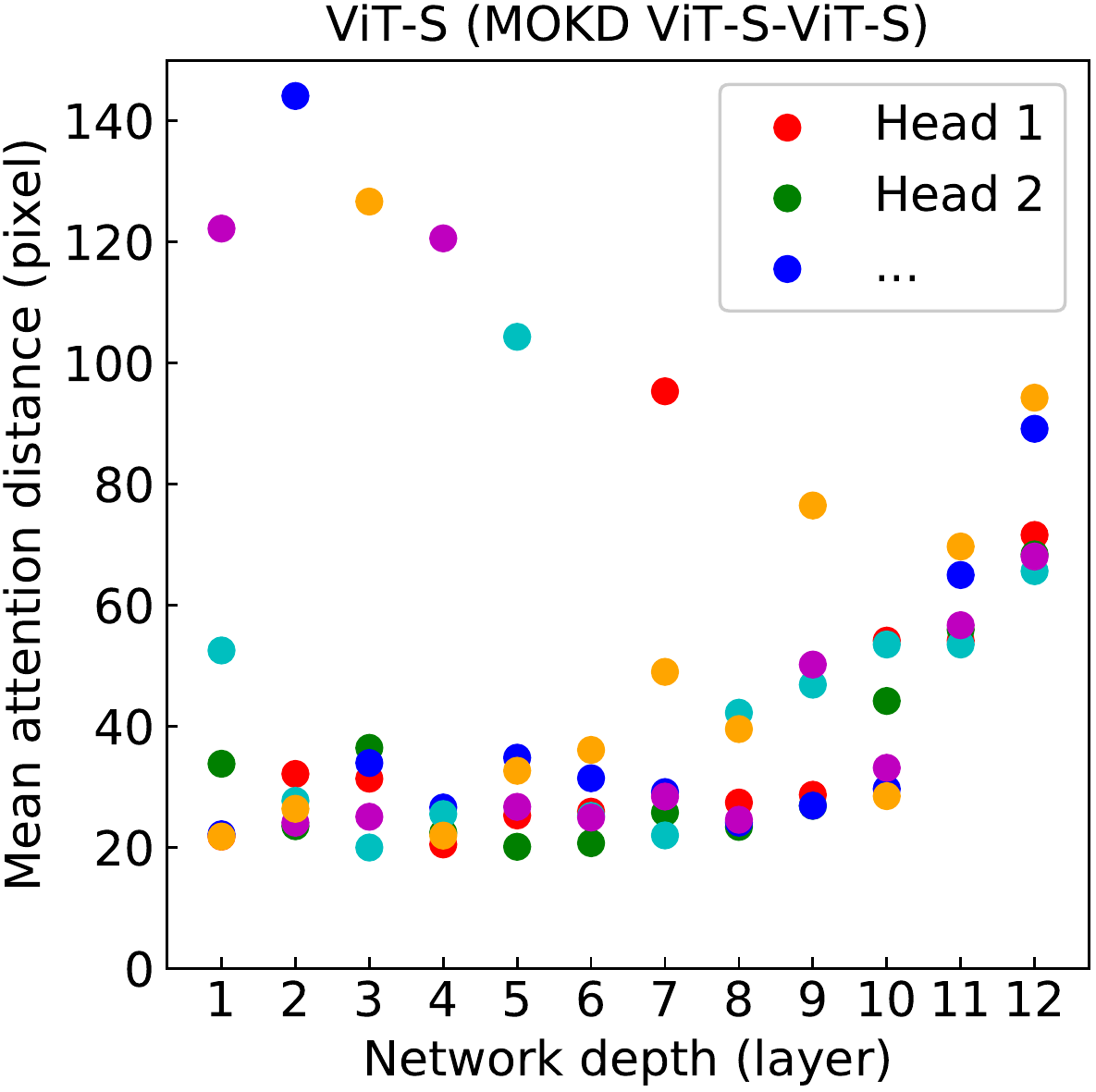}
    \caption{ViT-S (\method with ViT-S).}
    \label{fig_app_mad_c}
  \end{subfigure}
  \hfill
  \begin{subfigure}{0.23\linewidth}
    \includegraphics[width=1.6in]{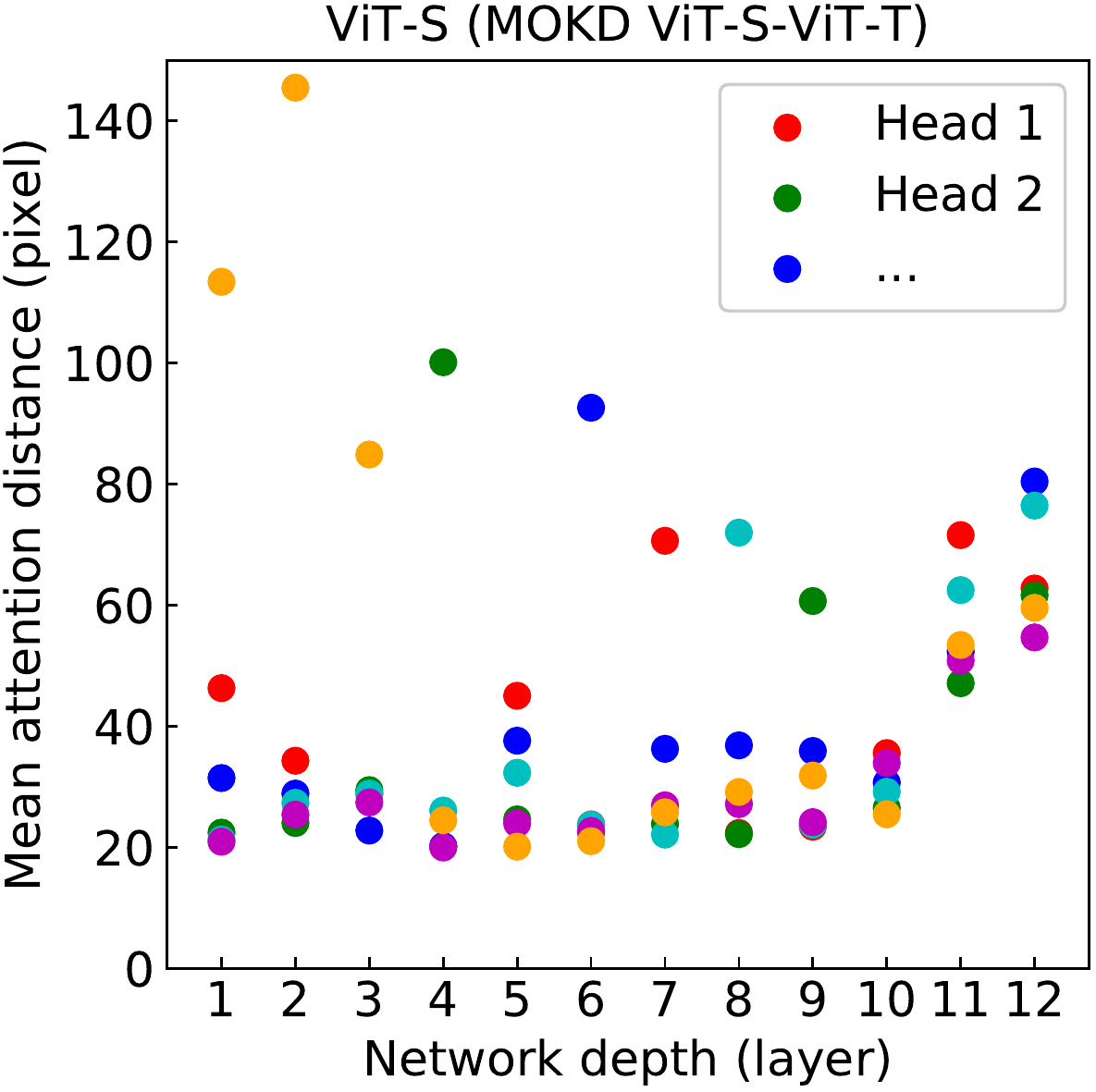}
    \caption{ViT-S (\method with ViT-T).}
    \label{fig_app_mad_d}
  \end{subfigure}
  
  \begin{subfigure}{0.25\linewidth}
    \includegraphics[width=1.6in]{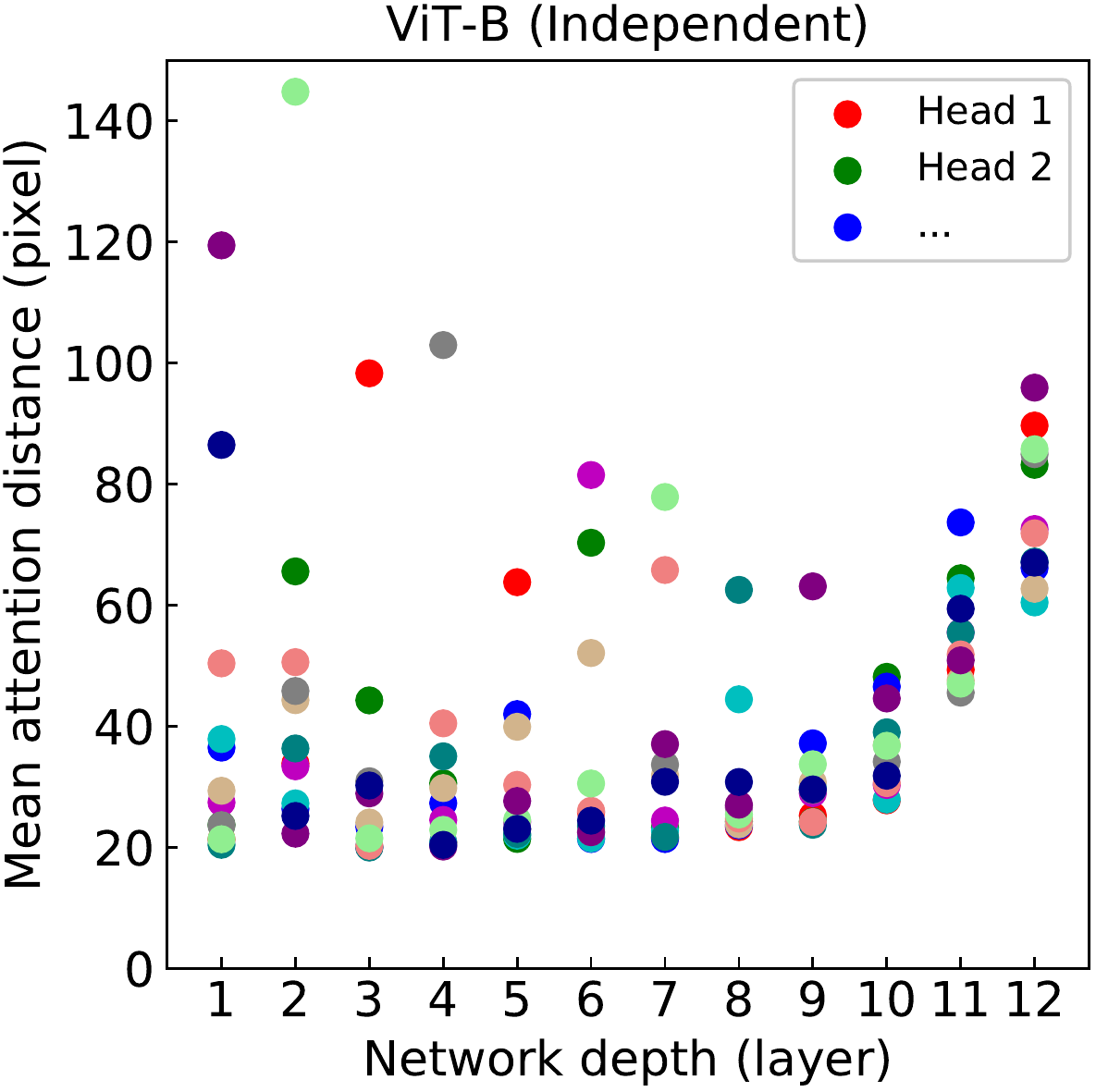}
    \caption{ViT-B (Independent).}
    \label{fig_app_mad_e}
  \end{subfigure}
  \begin{subfigure}{0.25\linewidth}
    \includegraphics[width=1.6in]{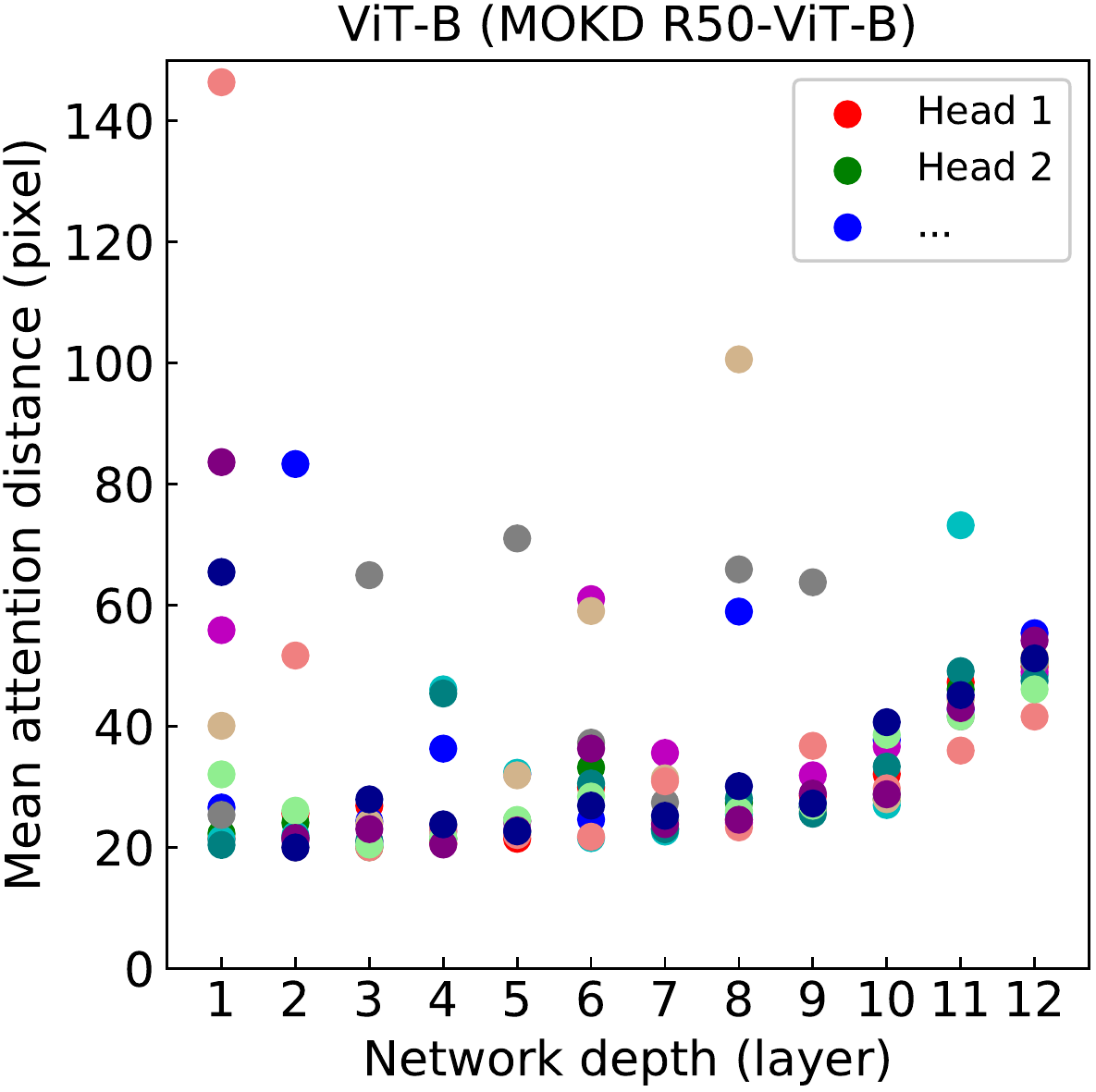}
    \caption{ViT-B (\method with R50).}
    \label{fig_app_mad_f}
  \end{subfigure}
  \begin{subfigure}{0.25\linewidth}
    \includegraphics[width=1.6in]{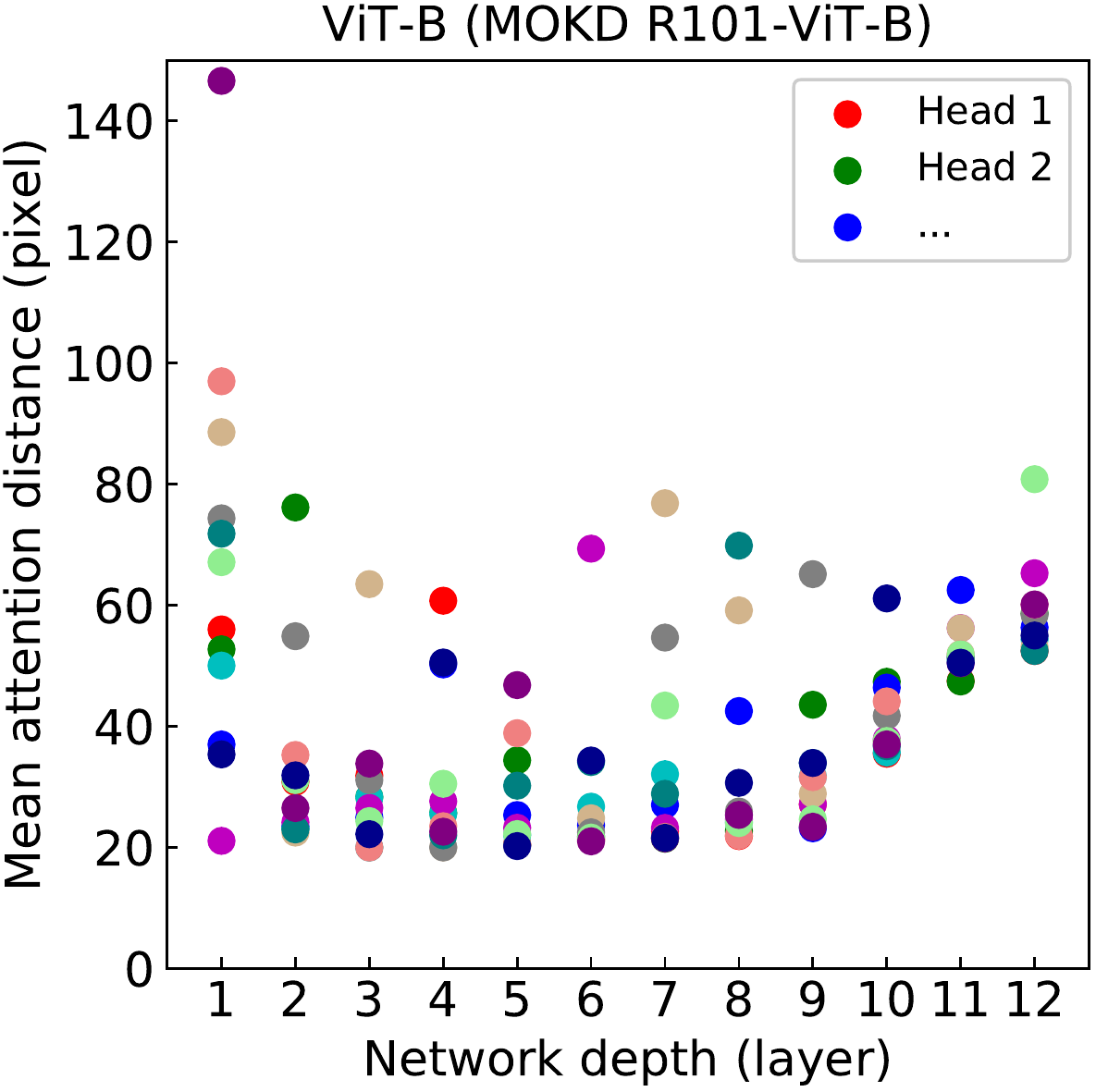}
    \caption{ViT-B (\method with R101).}
    \label{fig_app_mad_g}
  \end{subfigure}
  
  \caption{Mean attention distances \cite{vit_2021} of different models.}
  \label{fig_app_mad}
\end{figure*}

\subsection{Mean Attention Distances}
\label{subsec:sm_mad}

In Fig. \textcolor{red}{5}, we show the mean attention distances \cite{vit_2021} of ViT-S, R50, and R101 trained independently (by DINO) and trained by \method.
For ResNets, when trained with ViTs, i.e., R50-ViT-S, R5O-ViT-B, R101-ViT-S, and R101-ViT-B, the ResNet models trained by \method turns to be more ``global".
However, this phenomenon is not shown in two ResNet model pairs trained by \method, i.e., R50-R34, R50-R18, R101-R34, and R101-R18.
As shown in \cref{fig_app_mad}, we show more mean attention distances of ViTs.
Compared with the ViT models trained independently (as shown in \cref{fig_app_mad}(a)(e)), the mean attention distances on deep layers (layer 10-12) of the ViT models trained with ResNet models (as shown in \cref{fig_app_mad}(b)(f)(g)) decrease, which indicates that the ViT-S model trained by \method turns to be more ``local" on deep layers.
This phenomenon is not shown in two ViT models trained by \method, as shown in \cref{fig_app_mad}(c)(d).
These observations show that two heterogeneous models absorb knowledge from each other: ViT model learns more locality while CNN model learns more global information.

\section{More Experiments}
\label{sec_sm_results}

\subsection{Results on Other Convnets}
\label{subsec:sm_convnets}

We conduct experiments on EfficientNet-B0 \cite{efficientnet_2019} and MobileNet-v3-Large \cite{mobilenet_v3_2019}.
Following DisCo~\cite{disco_2022}, R50 and R101 are selected as the larger models.
We pre-train \method with 256 batch size for 100 epochs on ImageNet.
As shown in \cref{tab_ablation_otherconvnets}, \method brings consistent improvements over different model pairs.
It achieves the best performance for EfficientNet-B0 \cite{efficientnet_2019} and MobileNet-v3-Large \cite{mobilenet_v3_2019}.

\subsection{Influence of T-Head Depth}
\label{subsec:sm_thead}

T-Head is added for cross-attention feature search in \method.
As shown in \cref{tab_ablation_thead}, increasing T-Head depth (number of transformer blocks) improves the performance of \method.
However, extra computation costs brought by T-Head should be controlled.
Thus, T-Head depth should be small and set to 3 in this study.

\begin{table}[t]
\centering
\setlength{\tabcolsep}{0.4mm}
\scalebox{0.85}{
\begin{tabular}{l|cc|cc|cc|cc}
 \hline
Method & R50 & Eff-b0 & R50 & Mob-v3 & R101 & Eff-b0 & R101 & Mob-v3 \\ \hline
SEED~\cite{seed_2021} & 67.4 & 61.3 & 67.4 & 55.2 & 70.3 & 63.0 & 70.3 & 59.9 \\
ReKD~\cite{rekd_2022} & 67.6 & 63.4 & 67.6 & 56.7 & 69.7 & 65.0 & 69.7 & 59.6 \\
DisCo~\cite{disco_2022} & 67.4 & 66.5 & 67.4 & 64.4 & 69.1 & 68.9 & 69.1 & 65.7 \\
\rowcolor{Light} \textbf{\method} & 72.5 & \textbf{69.2} & 72.2 & \textbf{66.0} & 74.9 & \textbf{70.1} & 74.7 & \textbf{67.2} \\
\hline
\end{tabular}}
\caption{Results on other convnets.}
\label{tab_ablation_otherconvnets}
\end{table}

\begin{table}[t]
\centering
\begin{tabular}{c|c|c|c}
\toprule
Depth & 1 & 2 & 3 \\ \hline
R50, ViT-S & 87.4, 83.3 & 88.1, 84.1 & 88.3, 84.6 \\
\bottomrule
\end{tabular}
\caption{Influence of T-Head depth.}
\label{tab_ablation_thead}
\end{table}

\subsection{Training Time and Memory Requirement}
\label{subsec:sm_time}

We show the total training time and peak memory per GPU (“mem.”) when training ViT-S model pairs on an 8 V100 GPU machine.
We pre-train DINO~\cite{dino_2021} and \method with 256 batch size for 100 epochs on ImageNet.
From \cref{tab_ablation_time}, we can tell that the total memory requirement and training time of training two models independently via DINO~\cite{dino_2021} are comparable to those of \method.

\begin{table}[t]
\centering
\setlength{\tabcolsep}{2.8pt}
\begin{tabular}{llllll}
\hline
Method & Backbones & Time (h) & Mem.(G) & LP \\ \hline
DINO~\cite{dino_2021} & R50/ViT-S & 97+94 & 6.4+6.3 & 72.1/73.8 \\
\rowcolor{Light} \textbf{\method} & R50-ViT-S & 122 & 13.5 & 74.1/74.4 \\
\hline
\end{tabular}
\caption{Training Time and Memory Requirement.}
\label{tab_ablation_time}
\end{table}

\end{document}